\documentclass[10pt,twocolumn,letterpaper]{article}

\usepackage[pagenumbers]{cvpr}

\usepackage{tabularx}
\usepackage{multirow}
\usepackage{stfloats}
\usepackage{twemojis}

\newcommand{\glyphcenter}[1]{\raisebox{\dimexpr0.5\fontcharht\font`A-0.5\height+0.5\depth\relax}{#1}}
\newcommand{\glyphcell}[3][0.05\textwidth]{\glyphcenter{\includegraphics[width=#1]{figures/qualitative/#2/#3.pdf}}}
\newcommand{\glyphrow}[2]{#1
    & \glyphcell{#2}{0000}
    & \glyphcell{#2}{0001}
    & \glyphcell{#2}{0002}
    & \glyphcell{#2}{0003}
    & \glyphcell{#2}{0004}
    & \glyphcell{#2}{0005}
    & \glyphcell{#2}{0006}
    & \glyphcell{#2}{0007}
    & \glyphcell{#2}{0008}
    & \glyphcell{#2}{0009}
    & \glyphcell{#2}{0010}
    & \glyphcell{#2}{0011}
    & \glyphcell{#2}{0012}
    & \glyphcell{#2}{0013}
    & \glyphcell{#2}{0014}
    & \glyphcell{#2}{0015}
    \\}
\newcommand{\cp}[1]{& {\scriptsize U+#1}}

\newcommand{\glyphroweight}[2]{#1
    & \glyphcell[0.09\linewidth]{#2}{0000}
    & \glyphcell[0.09\linewidth]{#2}{0001}
    & \glyphcell[0.09\linewidth]{#2}{0002}
    & \glyphcell[0.09\linewidth]{#2}{0003}
    & \glyphcell[0.09\linewidth]{#2}{0004}
    & \glyphcell[0.09\linewidth]{#2}{0005}
    & \glyphcell[0.09\linewidth]{#2}{0006}
    & \glyphcell[0.09\linewidth]{#2}{0007}
    \\}

\definecolor{cvprblue}{rgb}{0.21,0.49,0.74}
\usepackage[pagebackref,breaklinks,colorlinks,allcolors=cvprblue]{hyperref}

\def\confName{CVPR}
\def\confYear{2027}

\title{FlowVecFont: Scalable Vector Font Generation via Latent Flow Matching}

\author{
    Takumu Fujioka$^{1}$ \qquad Gouhei Tanaka$^{1,2}$\\
    $^{1}$Department of Computer Science, Graduate School of Engineering,\\
    Nagoya Institute of Technology, Nagoya, 466-8555, Japan\\
    $^{2}$International Research Center for Neurointelligence,\\
    The University of Tokyo, Tokyo, 113-0033, Japan\\
    {\tt\small t.fujioka.494@stn.nitech.ac.jp, gtanaka@nitech.ac.jp}
}
\date{}

\begin{document}
\maketitle
\begin{abstract}
    Fonts are natively represented as vector outlines.
    Many existing studies on vector font generation have relied on specialized generation
    or refinement mechanisms,
    bitmap representations, or externally pretrained models to address the challenges
    of learning vector representations.
    These approaches often increase system complexity, incur additional computational
    and implementation costs, and introduce dependencies on external resources.
    We show that scaling a simple model enables high-quality pure vector font generation
    without external pretraining.
    We propose FlowVecFont, a scalable model for vector font generation
    based on latent flow matching and trained solely on vector outlines.
    A Transformer-based variational autoencoder maps glyph sequences
    into fixed-length continuous latent tokens, enabling standard conditional flow matching
    for vector glyph generation.
    The model supports conditioning on class labels or reference glyphs,
    with the latter enabling generalization to unseen characters and fonts.
    Experiments on 5M glyphs collected across Google Fonts, including non-Latin characters,
    show that generation quality improves with model scale across conditioning strategies.
    In a comparison on Latin glyphs, the 800M-parameter FlowVecFont outperforms
    DeepVecFont-v2 and DesigNet and achieves quality competitive with the pretrained
    27B-parameter VecGlypher.
    We also provide \iftoggle{cvprfinal}{TorchFont, a}{a}
    machine learning library for vector fonts.
\end{abstract}

\section{Introduction}


Developing fonts requires specialized expertise and considerable time.
In particular, developing CJK (Chinese, Japanese, and Korean) fonts can take several years
due to the large number of characters they contain.
These challenges have motivated growing interest in machine learning-based font generation
in recent years.


Font generation differs substantially from conventional image generation
in that fonts typically use vector rather than bitmap representations.
Unlike bitmap images, which represent visual content as a grid of pixels,
vector images represent shapes using lines and curves that define their outlines,
allowing fonts to be used at a wide range of sizes without loss of quality.

\begin{table}[t]
    \centering
    \caption{Drawing commands used to represent vector outlines.}
    \label{tab:drawing_commands}
    \small
    \renewcommand{\tabularxcolumn}[1]{m{#1}}
    \begin{tabularx}{\columnwidth}{@{}ccX@{}}
        \toprule
        Command & Arguments                                                                       & Description                                              \\
        \midrule
        MoveTo  & $x, y$                                                                          & Start a new subpath at $(x, y)$ without drawing.         \\
        \midrule
        LineTo  & $x, y$                                                                          & Draw a straight line from the current point to $(x, y)$. \\
        \midrule
        QuadTo  & \begin{tabular}{@{}c@{}}$c^x, c^y$ \\ $x, y$\end{tabular}
                & Draw a quadratic B\'ezier curve from the current point to $(x, y)$
        with control point $(c^x, c^y)$.
        Used in TrueType (\texttt{glyf}) outlines.                                                                                                           \\
        \midrule
        CurveTo & \begin{tabular}{@{}c@{}}$c_1^x, c_1^y$ \\ $c_2^x, c_2^y$ \\ $x, y$\end{tabular}
                & Draw a cubic B\'ezier curve from the current point to $(x, y)$
        with control points $(c_1^x, c_1^y)$ and $(c_2^x, c_2^y)$.
        Used in CFF/CFF2 outlines.                                                                                                                           \\
        \midrule
        Close   & $\varnothing$                                                                   & Close the current subpath with a straight line
        from the current point to its starting point.                                                                                                        \\
        \bottomrule
    \end{tabularx}
\end{table}

\begin{table*}[t]
    \centering
    \caption{Comparison of vector graphics and font generation methods.
        Approach denotes the model family:
        \texttwemoji{1f534} variational autoencoder (VAE) or encoder--decoder,
        \texttwemoji{1f7e2} diffusion or flow matching, and \texttwemoji{1f535} LLM.
        Font support indicates the number of distinct characters across training and evaluation:
        \texttwemoji{1f534} fewer than 1K, \texttwemoji{1f7e0} 1K to 10K, and \texttwemoji{1f7e2} 10K or more.
        Bitmap-free means no bitmap input or supervision, excluding evaluation.
        Pretraining-free means no external pretrained weights for training or generation,
        including loss networks but excluding evaluation-only metrics.
        Names beside crosses identify representative pretrained components, not an exhaustive list.}
    \label{tab:related_work}
    \small
    \begin{tabular*}{\textwidth}{@{\extracolsep{\fill}}lllccl@{}}
        \toprule
        Method & Approach                                 & Font support                        & Vector output      & Bitmap-free        & Pretraining-free                \\
        \midrule
        DeepSVG (2020)~\cite{carlier2020deepsvg}
        & \texttwemoji{1f534} VAE                  & \texttwemoji{1f534} Latin           & \texttwemoji{2705} & \texttwemoji{2705} & \texttwemoji{2705}              \\
        Aoki \& Aizawa (2022)~\cite{aoki2022svg}
        & \texttwemoji{1f534} Encoder--decoder     & \texttwemoji{1f534} Chinese         & \texttwemoji{2705} & \texttwemoji{2705} & \texttwemoji{2705}              \\
        DeepVecFont-v2 (2023)~\cite{wang2023deepvecfont}
        & \texttwemoji{1f534} VAE                  & \texttwemoji{1f534} Latin, Chinese  & \texttwemoji{2705} & \texttwemoji{274c} & \texttwemoji{274c} VGG-19 loss~\cite{simonyan2015very}  \\
        DualVector (2023)~\cite{liu2023dualvector}
        & \texttwemoji{1f534} VAE                  & \texttwemoji{1f534} Latin           & \texttwemoji{2705} & \texttwemoji{274c} & \texttwemoji{274c} LPIPS~\cite{zhang2018perceptual}        \\
        DesigNet (2026)~\cite{guijavaliente2026designet}
        & \texttwemoji{1f534} VAE                  & \texttwemoji{1f534} Latin, Chinese  & \texttwemoji{2705} & \texttwemoji{2705} & \texttwemoji{2705}              \\
        \midrule
        SVGFusion (2024)~\cite{xing2024svgfusion}
        & \texttwemoji{1f7e2} Latent diffusion     & \texttwemoji{274c}                  & \texttwemoji{2705} & \texttwemoji{274c} & \texttwemoji{274c} DINOv2~\cite{oquab2024dinov2}       \\
        VecFusion (2024)~\cite{thamizharasan2024vecfusion}
        & \texttwemoji{1f7e2} Diffusion            & \texttwemoji{1f534} Multiple        & \texttwemoji{2705} & \texttwemoji{274c} & \texttwemoji{2705}              \\
        Zhao et al. (2024)~\cite{zhao2024vector}
        & \texttwemoji{1f7e2} Diffusion            & \texttwemoji{1f534} Latin           & \texttwemoji{2705} & \texttwemoji{274c} & \texttwemoji{2705}              \\
        FontDiffuser (2024)~\cite{yang2024fontdiffuser}
        & \texttwemoji{1f7e2} Diffusion            & \texttwemoji{1f7e0} Chinese, Korean & \texttwemoji{274c} & \texttwemoji{274c} & \texttwemoji{274c} VGG-16 loss~\cite{simonyan2015very}  \\
        FM-Font (2026)~\cite{wang2026fmfont}
        & \texttwemoji{1f7e2} Flow matching        & \texttwemoji{1f7e0} Chinese         & \texttwemoji{274c} & \texttwemoji{274c} & \texttwemoji{274c} DINOv2~\cite{oquab2024dinov2}       \\
        \midrule
        OmniSVG (2025)~\cite{yang2025omnisvg}
        & \texttwemoji{1f535} Multimodal LLM       & \texttwemoji{274c}                  & \texttwemoji{2705} & \texttwemoji{274c} & \texttwemoji{274c} Qwen2.5-VL~\cite{bai2025qwen}   \\
        VecGlypher (2026)~\cite{huang2026vecglypher}
        & \texttwemoji{1f535} Multimodal LLM       & \texttwemoji{1f534} Latin           & \texttwemoji{2705} & \texttwemoji{274c} & \texttwemoji{274c} Gemma 3~\cite{gemmateam2025gemma}      \\
        VecFontLLM (2026)~\cite{yuan2026vecfontllm}
        & \texttwemoji{1f535} Multimodal LLM       & \texttwemoji{1f7e2} Chinese         & \texttwemoji{2705} & \texttwemoji{274c} & \texttwemoji{274c} StarCoder-1B~\cite{li2023starcoder} \\
        \midrule
        \textbf{FlowVecFont (ours)}
        & \texttwemoji{1f7e2} Latent flow matching & \texttwemoji{1f7e2} Multiple        & \texttwemoji{2705} & \texttwemoji{2705} & \texttwemoji{2705}              \\
        \bottomrule
    \end{tabular*}
\end{table*}


Table~\ref{tab:drawing_commands} lists the drawing commands used to construct font outlines.
Font outlines consist of variable-length sequences of these commands,
similar to handwriting trajectories.
Because vector graphics and bitmap images have fundamentally different data structures,
conventional approaches designed for bitmap images struggle to interpret and generate vector graphics.


Table~\ref{tab:related_work} compares representative methods for vector graphics and font generation.
Learning vector representations remains challenging, and many of these methods have
addressed this difficulty using specialized generation or refinement mechanisms,
bitmap representations, or external pretraining.
VecFusion~\cite{thamizharasan2024vecfusion}, a representative diffusion-based method,
fails to achieve satisfactory generation quality using vector representations alone.
Instead, it uses a multi-stage generation approach, first generating bitmap images
and then conditioning vector generation on them.
Language model-based methods such as VecGlypher~\cite{huang2026vecglypher}
and VecFontLLM~\cite{yuan2026vecfontllm} rely on knowledge acquired
through large-scale external pretraining as a workaround for the challenges of learning directly from vector representations.
These approaches can increase system complexity and introduce dependencies on external resources.
In particular, they require additional computation, but their benefits are often evaluated
by removing components without matching computational budgets.
This makes it difficult to determine how much of the improvement comes from
these design choices rather than the additional computation.

As an alternative, we show that scaling a simple model for pure vector font generation
can achieve competitive quality.
We propose FlowVecFont, a scalable model for vector font generation
based on latent flow matching~\cite{lipman2023flow},
without bitmap inputs, bitmap supervision, or externally pretrained weights.

Our main contributions are as follows:
\begin{itemize}
    \item We propose FlowVecFont, which applies flow matching to fixed-length glyph latents
          learned by a Transformer VAE, decoupling flow computation from outline length.
    \item We demonstrate generalization to unseen fonts and characters
          by conditioning on vector reference glyphs
          for both font style and character content.
    \item Experiments on Google Fonts~\cite{googlefonts} show that generation quality
          consistently improves with model scale across conditioning strategies.
          Our 800M model is competitive with a pretrained 27B model on Latin glyphs.
    \item We provide \iftoggle{cvprfinal}{TorchFont}{[name anonymized for review]},
          a machine learning library for vector fonts\iftoggle{cvprfinal}
          { (\href{https://github.com/torchfont/torchfont}{https://github.com/torchfont/torchfont})}{}.
\end{itemize}

\section{Related Work}

\subsection{Vector Graphics and Font Generation}

\paragraph{VAE and Encoder--Decoder Models.}


DeepSVG~\cite{carlier2020deepsvg}, an early method for vector graphics and font generation,
introduces a hierarchical Transformer VAE that models paths and drawing commands at separate levels.
Subsequent methods build on DeepSVG's vector representation or model design,
including Aoki and Aizawa~\cite{aoki2022svg} and DesigNet~\cite{guijavaliente2026designet}.
Aoki and Aizawa generate Chinese vector fonts from content and style references,
using a Transformer encoder--decoder and geometric losses.
DesigNet extends hierarchical VAE modeling
with differentiable refinement of curve continuity and axis alignment,
incorporating geometric constraints used in professional typeface design.


Another line of work, including DeepVecFont~\cite{wang2021deepvecfont},
DeepVecFont-v2~\cite{wang2023deepvecfont}, and DualVector~\cite{liu2023dualvector},
proposes using bitmap representations for vector font generation.
DeepVecFont-v2 combines image and sequence features with Transformer-based outline refinement.
DualVector learns positive and negative paths from glyph images
and constructs contours through Boolean operations.
These studies report improved generation quality through the use of bitmap information.

\paragraph{Diffusion and Flow Matching Models.}


VecFusion~\cite{thamizharasan2024vecfusion} and Zhao et al.~\cite{zhao2024vector}
apply diffusion models to vector font generation.
VecFusion generates bitmap glyphs and control point fields first,
then uses them to guide vector generation.
Zhao et al. jointly model vector and bitmap diffusion,
exchanging features through cross-attention at each denoising step.
Both methods rely on bitmap information during training and generation.

Related methods address either general SVG generation or bitmap font generation.
SVGFusion~\cite{xing2024svgfusion} applies latent diffusion to general SVG graphics rather than font glyphs,
but learns its latent representation using bitmap features from a pretrained DINOv2 encoder.
FontDiffuser~\cite{yang2024fontdiffuser} and FM-Font~\cite{wang2026fmfont}
apply diffusion and flow matching, respectively, to bitmap font generation.
Neither produces vector outlines.

\paragraph{Large Language Models.}


SVG is an XML-based text format that describes graphics through elements and attributes,
allowing SVG synthesis to be formulated as a code generation task for language models.
OmniSVG~\cite{yang2025omnisvg} adapts a pretrained vision-language model to general SVG generation
by fine-tuning Qwen2.5-VL to predict discrete tokens representing drawing commands,
coordinates, and colors under text or image conditioning.

This approach has subsequently been extended to vector font generation
by VecGlypher~\cite{huang2026vecglypher} and VecFontLLM~\cite{yuan2026vecfontllm}.
VecGlypher fine-tunes Gemma 3 to autoregressively generate SVG outlines
conditioned on textual style descriptions or reference glyph images.
VecFontLLM uses a multi-stage generation process based on StarCoder-1B,
first predicting the anchor layout and then completing B\'ezier control points
to form complex Chinese glyphs.

\subsection{Generative Models}

Denoising diffusion models~\cite{ho2020denoising} generate samples
by learning to reverse a gradual noising process.
Flow matching~\cite{lipman2023flow} shares the idea of progressively transforming noise into data,
but directly learns a velocity field along prescribed probability paths,
including those associated with diffusion processes.
Samples are generated by integrating the learned velocity field.
Rectified Flow~\cite{liu2023rectified} is a form of flow matching
that uses straight-line interpolation between noise and data samples.

Latent diffusion and latent flow matching separate representation learning from generation,
modeling the data distribution in a learned latent space~\cite{rombach2022latent}.
Transformer-based models such as DiT~\cite{peebles2023scalable}
and SiT~\cite{ma2024sit} demonstrate the scalability of this approach.
Our method applies this framework to vector drawing command sequences.

\section{Method}

\begin{figure*}[t]
    \centering
    \captionsetup[subfigure]{justification=centering}
    \def\architecturescale{0.65}
    \begin{subfigure}[c]{0.315\textwidth}
        \centering
        \includegraphics[scale=\architecturescale]{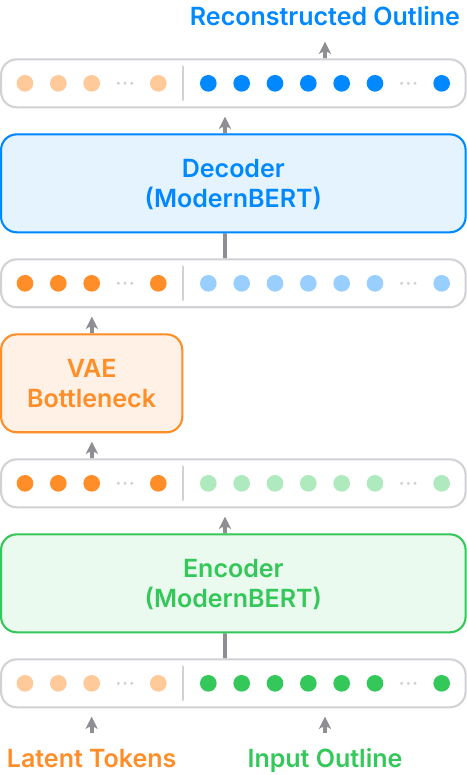}
        \caption{VAE}
        \label{fig:vae_architecture}
    \end{subfigure}
    \hfill
    \begin{minipage}{0.66\textwidth}
        \centering
        \begin{subfigure}{0.46\linewidth}
            \centering
            \includegraphics[scale=\architecturescale]{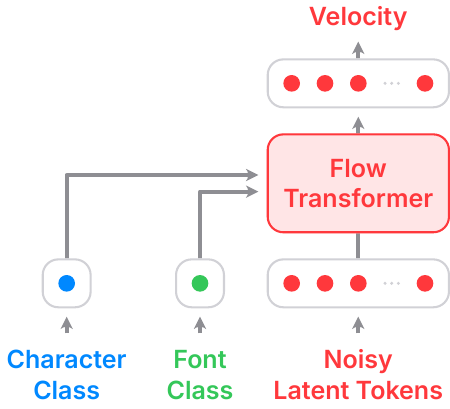}
            \caption{Flow matching with\\character and font classes (CLS)}
            \label{fig:flow_class_conditioning}
        \end{subfigure}
        \hfill
        \begin{subfigure}{0.51\linewidth}
            \centering
            \includegraphics[scale=\architecturescale]{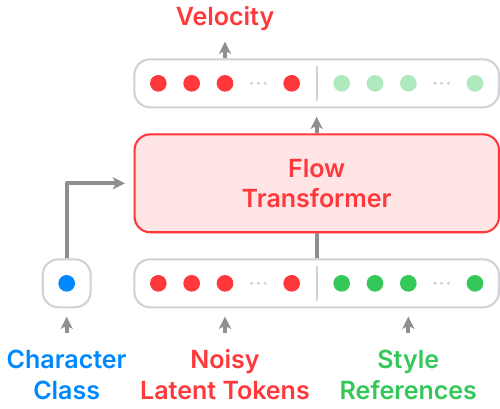}
            \caption{Flow matching with\\character class and style references (STR)}
            \label{fig:flow_character_class_style_references}
        \end{subfigure}

        \bigskip
        \begin{subfigure}{0.56\linewidth}
            \centering
            \includegraphics[scale=\architecturescale]{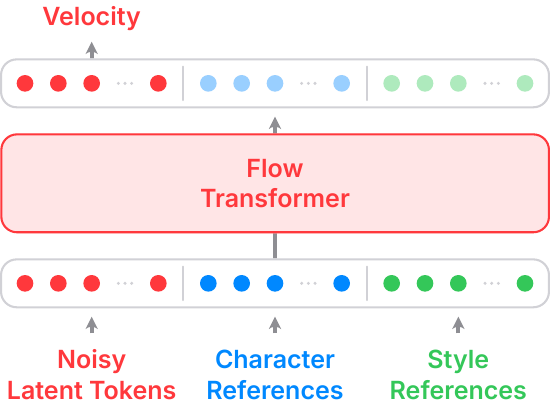}
            \caption{Flow matching with\\character and style references (REF)}
            \label{fig:flow_character_and_style_references}
        \end{subfigure}
    \end{minipage}
    \caption{FlowVecFont architectures.
        \subref{fig:vae_architecture} The VAE encodes and reconstructs vector outlines.
        \subref{fig:flow_class_conditioning}--\subref{fig:flow_character_and_style_references}
        Latent flow matching generates glyph representations
        using different conditioning strategies.}
    \label{fig:flow_conditioning}
\end{figure*}

\subsection{Overview}

FlowVecFont generates vector glyphs using latent flow matching,
conditioned on character content and font style.


Direct generative modeling of vector outlines requires handling both discrete
command types and continuous coordinates, motivating specialized representations.
VecFusion~\cite{thamizharasan2024vecfusion} encodes path membership and grid cell
coordinates as binary bits, which are diffused as continuous values and discretized
after generation.
Zhao et al.~\cite{zhao2024vector} encode command types as two-dimensional numerical
codes and jointly diffuse these codes and geometric parameters.
Moreover, when diffusion operates directly on outline elements, increasing the
maximum sequence length increases the computational cost of each denoising step.


FlowVecFont addresses these challenges using a Transformer VAE and latent
flow matching.
The VAE encoder aggregates outline information into a fixed number of learnable
latent tokens through self-attention, compressing variable-length command sequences
into fixed-size representations.
It jointly encodes discrete command types and continuous coordinates into a learned
continuous latent space.
We then train a flow matching model in this space, conditioned on character content
and font style.
At inference time, the flow model transforms Gaussian noise into a glyph latent,
which the VAE decoder converts into a drawing command sequence.


This design allows the flow model to operate entirely on fixed-size continuous
representations, without directly modeling discrete command types and coordinates.
With a fixed number of latent tokens, the supported outline sequence length can be
increased on the VAE side without changing the flow architecture or its per-step
computational cost.
This formulation allows vector font generation to use essentially the same
generative architecture as latent flow matching for images.

\subsection{Preprocessing}


We first exclude glyphs with outline sequences longer than the maximum length $L$.
Fonts use different coordinate scales, so we then divide outline coordinates by
each font's units per em to obtain a common scale.
The resulting coordinates lie approximately in $[0, 1]$.


We represent glyph outlines using six command types:
MoveTo, LineTo, CurveTo, Close, End, and Pad.
Quadratic B\'ezier curves (QuadTo) are converted to cubic B\'ezier curves (CurveTo).
End terminates the sequence, and Pad fills unused positions for batching.


We define the command sequence for each glyph as follows:
\begin{equation}
    \mathcal{G}
    = \bigl[(t_1, \mathbf{p}_1), \ldots, (t_N, \mathbf{p}_N)\bigr],
\end{equation}
where $N$ is the sequence length, $t_i$ denotes the command type, and
$\mathbf{p}_i$ is its coordinate vector.
The coordinate vector $\mathbf{p}_i \in \mathbb{R}^{6}$ is defined as follows (see Table~\ref{tab:drawing_commands}):
\begin{equation}
    \mathbf{p}_i = (c_{i,1}^{x}, c_{i,1}^{y},
    c_{i,2}^{x}, c_{i,2}^{y}, x_i, y_i)^{\top}.
\end{equation}
It contains the coordinates of up to two control points and an endpoint.
Unused coordinate entries are set to zero.


We then apply outline normalization and data augmentation.
These transforms include contour winding adjustment, subpath starting point
selection, subpath reordering, geometric transformations, overlap removal,
segment splitting, and subpath dropout.
The choice of transforms, their strengths, and their use for normalization or
random data augmentation depend on the model type, conditioning scheme, and
training or evaluation setting.


These transforms can increase sequence length.
When a sequence exceeds the maximum length $L$, we remove complete subpaths until it fits.
Finally, we pad the sequence with Pad tokens to the maximum length for batching.

\subsection{Variational Autoencoder}

To learn a latent representation for generation, we train a VAE to reconstruct
the input command sequence in its original path order.
Unlike DeepSVG~\cite{carlier2020deepsvg},
we do not impose invariance to path ordering.
As shown in Figure~\ref{fig:vae_architecture}, the encoder and decoder follow
the 1D image tokenizer design of TiTok~\cite{yu2024image}.
Both use the ModernBERT architecture~\cite{warner2025modernbert} and are trained from scratch.
The decoder is non-autoregressive.

\begin{figure}[t]
    \centering
    \includegraphics[width=\columnwidth]{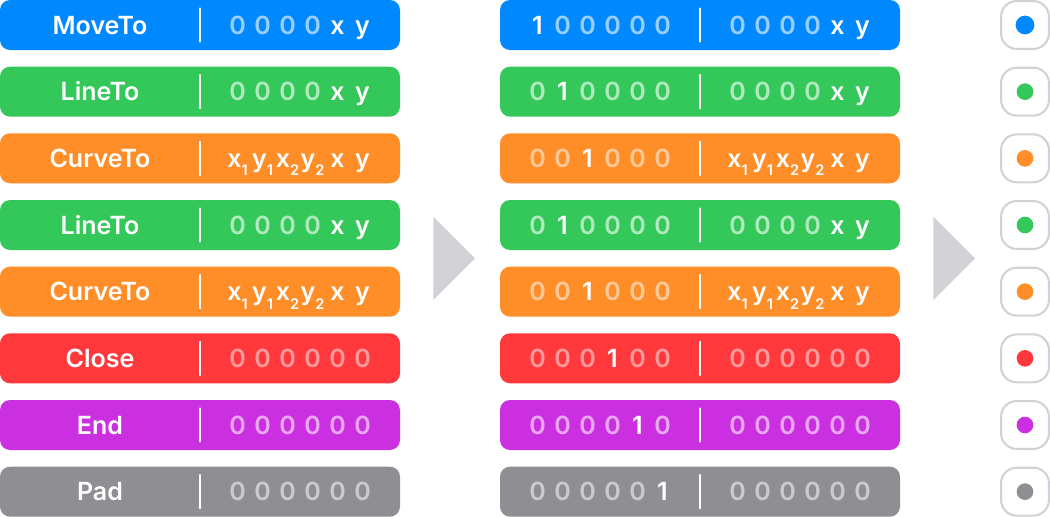}
    \caption{Drawing command embedding.
        Command types are one-hot encoded and embedded, while coordinates are embedded
        through a learnable linear layer.
        The two embeddings are added to form a command token.}
    \label{fig:data_representation}
\end{figure}

As illustrated in Figure~\ref{fig:data_representation}, we embed command types and
coordinates separately.
We define the resulting $d$-dimensional command token as follows:
\begin{equation}
    \mathbf{e}_i = \mathbf{W}_{\mathrm{type}}\operatorname{onehot}(t_i)
    + \mathbf{W}_{\mathrm{coord}}\mathbf{p}_i,
\end{equation}
where $\operatorname{onehot}(t_i) \in \mathbb{R}^{6}$ is the one-hot encoding of $t_i$, and
$\mathbf{W}_{\mathrm{type}} \in \mathbb{R}^{d \times 6}$ and
$\mathbf{W}_{\mathrm{coord}} \in \mathbb{R}^{d \times 6}$ are learnable matrices.
Coordinates remain continuous and are not quantized.

The encoder compresses the command sequence into $K$ latent tokens.
We concatenate $K$ learnable tokens with the command embeddings and process the
resulting sequence using self-attention.
The outputs at the latent token positions parameterize the latent representation
$\mathbf{Z} \in \mathbb{R}^{K \times d_z}$, where $d_z$ is the dimension per token.

The decoder reconstructs $L$ command tokens from the latent representation through
self-attention, then maps them to command type logits and six-dimensional
coordinates through an unembedding layer.

We define the VAE training objective as follows:
\begin{equation}
    \mathcal{L}_{\mathrm{VAE}}
    = \lambda_{\mathrm{type}}\mathcal{L}_{\mathrm{type}}
    + \lambda_{\mathrm{coord}}\mathcal{L}_{\mathrm{coord}}
    + \lambda_{\mathrm{KL}}\mathcal{L}_{\mathrm{KL}},
\end{equation}
where $\mathcal{L}_{\mathrm{type}}$ is the command type cross-entropy loss over
non-Pad positions, $\mathcal{L}_{\mathrm{coord}}$ is the coordinate loss defined below, and $\mathcal{L}_{\mathrm{KL}}$ is the standard VAE KL
regularization loss.
The weights $\lambda_{\mathrm{type}}$, $\lambda_{\mathrm{coord}}$, and
$\lambda_{\mathrm{KL}}$ control the contributions of the three losses.

Prior work such as DesigNet~\cite{guijavaliente2026designet} uses mean squared error
for coordinate reconstruction.
Inspired by the use of L1 reconstruction loss in raster image
generation~\cite{rombach2022latent}, we use L1 loss for coordinates to prevent
model collapse.
For each glyph, we define the masked L1 coordinate loss as follows:
\begin{equation}
    \mathcal{L}_{\mathrm{coord}}
    = \sum_{i=1}^{L}
    \left\|\mathbf{m}(t_i) \odot
    (\hat{\mathbf{p}}_i - \mathbf{p}_i)\right\|_1,
\end{equation}
where $\hat{\mathbf{p}}_i$ is the predicted coordinate vector and $\mathbf{m}(t_i)$ is
a mask to select the coordinates used by the target command.
We define this mask as follows:
\begin{equation}
    \mathbf{m}(t) =
    \begin{cases}
        (0,0,0,0,1,1)^{\top} & t \in \{\text{MoveTo},\text{LineTo}\},        \\
        (1,1,1,1,1,1)^{\top} & t = \text{CurveTo},                           \\
        (0,0,0,0,0,0)^{\top} & t \in \{\text{Close},\text{End},\text{Pad}\}.
    \end{cases}\!
\end{equation}
Both reconstruction losses are summed within each glyph and averaged over the batch.

\subsection{Flow Matching}


We adopt rectified flow~\cite{liu2023rectified} for conditional
flow matching~\cite{lipman2023flow} in the latent space of the trained VAE.
For each training glyph $\mathcal{G}$, we obtain its latent representation
$\mathbf{Z}_0 \in \mathbb{R}^{K \times d_z}$ using the VAE encoder.
We define the noisy latent by linear interpolation as follows:
\begin{equation}
    \mathbf{Z}_t = (1-t)\mathbf{Z}_0 + t\boldsymbol{\epsilon},
    \qquad
    \boldsymbol{\epsilon} \sim \mathcal{N}(\mathbf{0}, \mathbf{I}),
\end{equation}
where $t \sim \mathcal{U}(0,1)$, with $t=0$ corresponding to data
and $t=1$ to noise.
For each sampled pair $(\mathbf{Z}_0, \boldsymbol{\epsilon})$,
the target velocity $\mathbf{u}$ is the rate of change of $\mathbf{Z}_t$
with time, given as follows:
\begin{equation}
    \mathbf{u}
    = \frac{\mathrm{d}\mathbf{Z}_t}{\mathrm{d}t}
    = \boldsymbol{\epsilon}-\mathbf{Z}_0.
\end{equation}


We train the velocity field $v_{\theta}(\mathbf{Z}_t,t,c)$ to predict
$\mathbf{u}$, where $\theta$ denotes the learnable model parameters
and $c$ denotes the character content and font style conditions.
We parameterize $v_{\theta}$ using a Transformer,
following DiT~\cite{peebles2023scalable}.
We define the mean squared error objective as follows:
\begin{equation}
    \mathcal{L}_{\mathrm{FM}}
    =
    \mathbb{E}
    \left[
        \left\|
        v_{\theta}(\mathbf{Z}_t,t,c) - \mathbf{u}
        \right\|^2
        \right].
\end{equation}

At inference time, we initialize
$\mathbf{Z}_1 \sim \mathcal{N}(\mathbf{0},\mathbf{I})$.
The corresponding ordinary differential equation is defined as follows:
\begin{equation}
    \frac{\mathrm{d}\mathbf{Z}_t}{\mathrm{d}t}
    = v_{\theta}(\mathbf{Z}_t,t,c).
\end{equation}
We solve this equation backward in time to $t=0$ using the Euler method,
transporting the initial noise toward the conditional latent data distribution.
The VAE decoder converts the resulting latent into a drawing
command sequence.

We use classifier-free guidance (CFG)~\cite{ho2021classifier}, randomly
dropping both character and style conditions during training.
During generation, we combine conditional and unconditional velocity predictions
to control the strength of conditioning.

\subsection{Conditioning Strategies}

As shown in Figures~\ref{fig:flow_class_conditioning}--\ref{fig:flow_character_and_style_references},
we compare three conditioning strategies for character content and font style:
character and font classes (CLS),
character class and style references (STR),
and character and style references (REF).
Style references are glyphs from the target font, while character references
are glyphs of the target character from other fonts.

For class conditioning, we inject learnable class embeddings through
AdaLN-Zero, following DiT~\cite{peebles2023scalable}.

For reference conditioning, we encode each vector reference glyph using
the same trained VAE encoder used for the target glyph.
Reference latents remain fixed and noise-free during sampling.
We project the reference latents to the Transformer dimension and concatenate
them with the noisy target tokens as in-context conditioning.
The combined sequence is processed jointly through self-attention.
Positional embeddings identify latent token positions, while learned role
and reference embeddings distinguish the target, style references,
character references, and individual references within each group.
Only the outputs at the target token positions are used to predict velocity.

Class conditioning relies on embeddings learned for individual class IDs
and cannot directly handle unseen fonts or characters in this formulation.
Reference conditioning instead specifies style and content through glyph outlines,
without requiring a learned embedding for each font or character.
Given suitable references, STR can therefore handle unseen fonts in principle,
while REF can handle both unseen fonts and unseen characters.

\section{Experiments}

\subsection{Implementation Details}

The VAE encoder and decoder use the model size configuration
of ViT-Tiny~\cite{zhai2022scaling}.
For each conditioning strategy, we train flow models using the model size
configurations of DiT-S, DiT-B, DiT-L, and DiT-XL~\cite{peebles2023scalable}.
We name each variant by its conditioning strategy and model size,
e.g., FlowVecFont-CLS/B or FVF-CLS/B.

The maximum command sequence length is $L=256$.
The VAE uses $K=32$ latent tokens with $d_z=16$ dimensions per token.
The VAE loss weights are
$\lambda_{\mathrm{type}}=1$,
$\lambda_{\mathrm{coord}}=3$, and
$\lambda_{\mathrm{KL}}=10^{-4}$.

We train the VAE and flow models using AdamW with a constant learning rate
of $10^{-4}$ and BF16 mixed precision.
The VAE is trained for 400K steps, followed by 1M steps for each flow model.
The batch sizes are 64 for the VAE and 256, 64, and 32 for
CLS, STR, and REF, respectively.
We maintain an exponential moving average of model weights with a decay
of 0.9999.

For reference conditioning, we use four glyphs per reference type.
References are randomly sampled without replacement from each group,
excluding the target glyph.
For CFG, conditions are dropped with probability 0.1 during training,
and the guidance scale is 3.0 at inference time, with 32 Euler steps.

\subsection{Dataset and Evaluation}

We use vector glyphs from Google Fonts~\cite{googlefonts},
treating each font-face--codepoint pair as a sample.
We collect TTF files distributed under the Apache License,
SIL Open Font License, and Ubuntu Font Licence, excluding Adobe Blank.
We collect outline glyphs mapped to codepoints, excluding unencoded ligatures and alternates,
without restricting the character set to a particular language or script.
After excluding glyphs longer than $L$, the dataset contains approximately 4.9M samples from 3.8K font faces
and 109K distinct codepoints.
Each variable font is counted as a single face, regardless of its variation-axis settings.

We adopt the seen/unseen partitioning scheme used in bitmap font generation,
such as FontDiffuser~\cite{yang2024fontdiffuser}, for vector font generation.
Based on whether the target font and character were observed during training,
we define four groups:
seen fonts and seen characters (SFSC), seen fonts and unseen characters (SFUC),
unseen fonts and seen characters (UFSC), and unseen fonts and unseen characters (UFUC).

Font faces and character codepoints are independently partitioned at random,
with 95\% designated as seen and 5\% as unseen.
The SFSC group is further divided into training, validation,
and test subsets with proportions of 98\%, 1\%, and 1\%, respectively.
In the results, SFSC denotes the test subset of this group.
All models use the same partitions.

Table~\ref{tab:evaluation_splits} summarizes the evaluation splits
and reference pools for CLS, STR, and REF.
Targets with insufficient references are omitted.

We evaluate glyphs using symmetric Chamfer distance (CD)~\cite{barrow1977parametric},
learned perceptual image patch similarity (LPIPS)~\cite{zhang2018perceptual},
Fr\'echet Inception Distance (FID)~\cite{heusel2017gans}, and the evaluation failure rate.
Evaluation failures include empty outlines and certain errors during evaluation.
Quality metrics exclude failed samples and their targets.

\begin{table}[t]
    \centering
    \caption{Evaluation splits, reference pools, and sample counts by conditioning strategy.
        Train denotes the SFSC training subset.}
    \label{tab:evaluation_splits}
    \small
    \begin{tabular*}{\columnwidth}{@{\extracolsep{\fill}}llllr@{}}
        \toprule
        Strategy & Target & Style refs. & Char. refs. & Samples \\
        \midrule
        CLS & SFSC & --- & --- & 44.5K \\
        \midrule
        STR & SFSC & Train & --- & 44.5K \\
        STR & UFSC & UFSC & --- & 211.8K \\
        \midrule
        REF & SFSC & Train & Train & 43.9K \\
        REF & SFUC & Train & SFUC & 219.5K \\
        REF & UFSC & UFSC & Train & 209.6K \\
        REF & UFUC & UFSC & SFUC & 10.3K \\
        \bottomrule
    \end{tabular*}
\end{table}

\subsection{Conditioning Strategies and Model Scaling}

\begin{figure}[t]
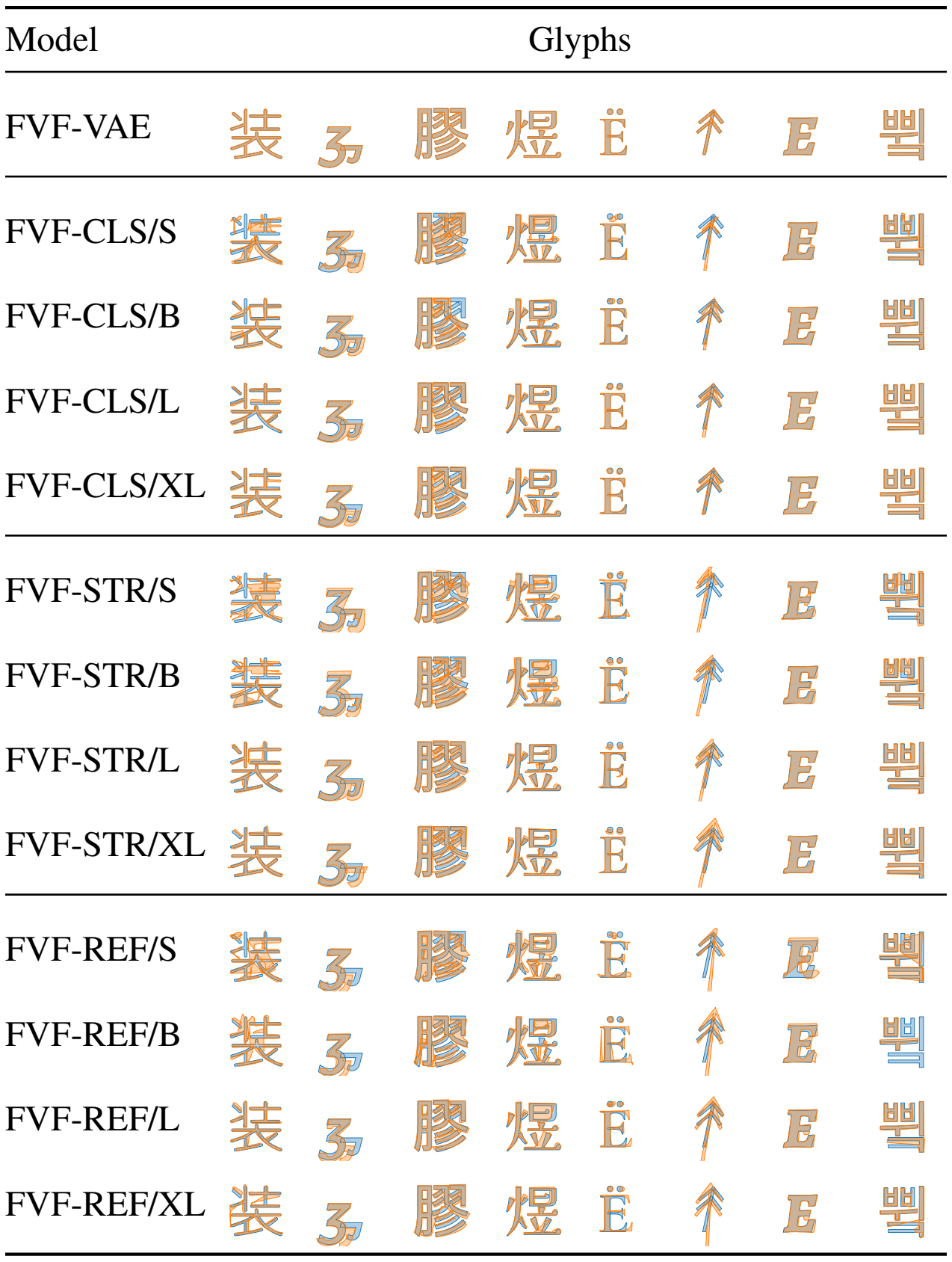

    \centering
    \small
    \setlength{\tabcolsep}{0pt}
    \begin{tabular*}{\linewidth}{@{\extracolsep{\fill}}l*{8}{c}@{}}
        \toprule
        Model & \multicolumn{8}{c@{}}{Glyphs} \\
        \midrule
        \glyphroweight{FVF-VAE}{vae-sfsc}
        \midrule
        \glyphroweight{FVF-CLS/S}{cls-s-sfsc}
        \glyphroweight{FVF-CLS/B}{cls-b-sfsc}
        \glyphroweight{FVF-CLS/L}{cls-l-sfsc}
        \glyphroweight{FVF-CLS/XL}{cls-xl-sfsc}
        \midrule
        \glyphroweight{FVF-STR/S}{str-s-sfsc}
        \glyphroweight{FVF-STR/B}{str-b-sfsc}
        \glyphroweight{FVF-STR/L}{str-l-sfsc}
        \glyphroweight{FVF-STR/XL}{str-xl-sfsc}
        \midrule
        \glyphroweight{FVF-REF/S}{ref-s-sfsc}
        \glyphroweight{FVF-REF/B}{ref-b-sfsc}
        \glyphroweight{FVF-REF/L}{ref-l-sfsc}
        \glyphroweight{FVF-REF/XL}{ref-xl-sfsc}
        \bottomrule
    \end{tabular*}
    \caption{VAE reconstructions and generated glyphs on SFSC.
        Ground-truth outlines (blue) are overlaid with reconstructions or generated glyphs (orange).
        Columns share the same targets; samples are not cherry-picked.}
    \label{fig:conditioning_qualitative}
\end{figure}


Figure~\ref{fig:conditioning_qualitative} compares VAE reconstructions and generated glyphs
for each conditioning strategy and model size on SFSC.
The reconstructions almost perfectly overlap the target outlines.
Smaller models exhibit visible artifacts, particularly in complex glyphs such as
Chinese characters with many strokes, whereas larger models generally produce cleaner outlines.
REF can generate well-formed glyphs whose outlines differ from the targets.

As shown in Table~\ref{tab:conditioning_results}, the VAE achieves a CD of 0.0029
and an LPIPS of 0.0045, with no evaluation failures.
These reconstruction errors are substantially lower than the generation errors,
indicating that the VAE provides a suitable representation for latent generation.
For generation, all four metrics consistently improve as model size increases
from S to XL for each conditioning strategy.
CLS performs best across all four metrics at each model size on SFSC.
However, the strategies use different batch sizes, so the results do not necessarily
reflect the effect of conditioning alone.

\begin{figure}[t]
    \centering
    \includegraphics[width=\columnwidth]{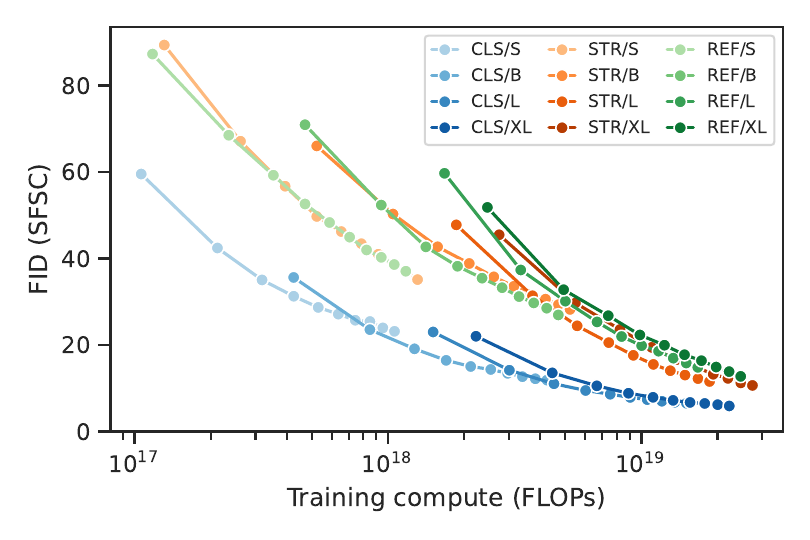}
    \caption{FID on SFSC versus estimated flow-model training compute.
        Each curve tracks a model across checkpoints saved every 100K training steps.}
    \label{fig:scaling_fid}
\end{figure}

Figure~\ref{fig:scaling_fid} shows FID against estimated flow-model training compute.
We estimate training FLOPs following DiT~\cite{peebles2023scalable}.
FID decreases with training compute for every model and conditioning strategy,
and larger models reach lower FID at the end of training.
This demonstrates that generation quality improves with model and compute scaling.
For CLS, B overtakes S and L overtakes B as training compute increases.
This indicates that larger models can use training compute more efficiently,
as observed for DiT.

\subsection{Generalization to Unseen Fonts and Characters}

\begin{figure}[t]
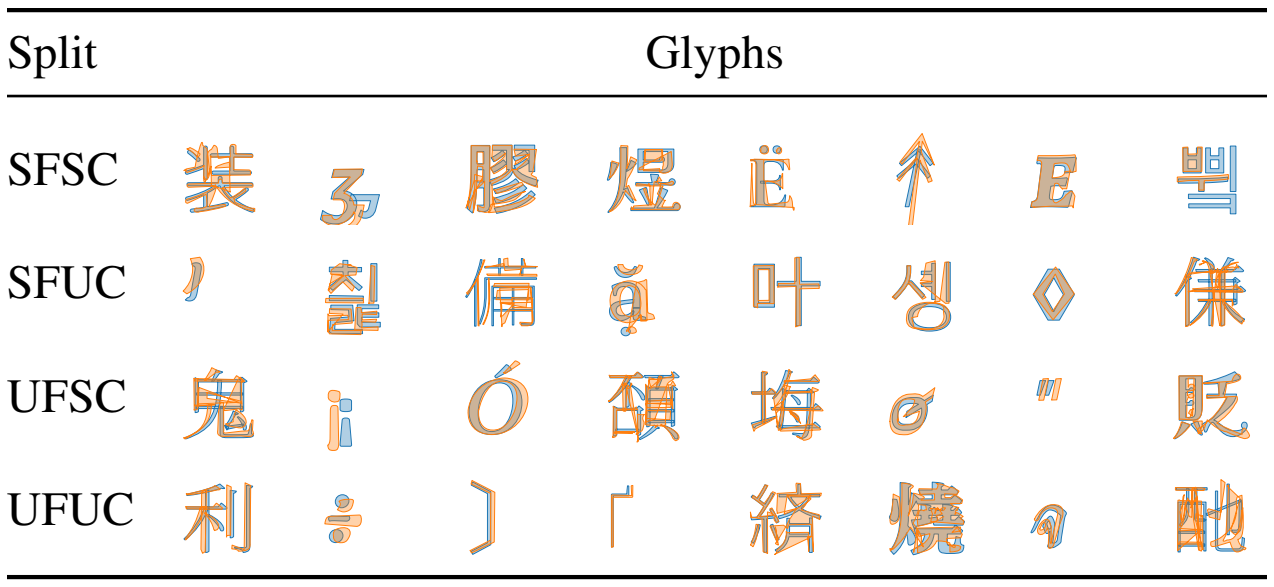

    \centering
    \small
    \setlength{\tabcolsep}{0pt}
    \begin{tabular*}{\linewidth}{@{\extracolsep{\fill}}l*{8}{c}@{}}
        \toprule
        Split & \multicolumn{8}{c@{}}{Glyphs} \\
        \midrule
        \glyphroweight{SFSC}{ref-b-sfsc}
        \glyphroweight{SFUC}{ref-b-sfuc}
        \glyphroweight{UFSC}{ref-b-ufsc}
        \glyphroweight{UFUC}{ref-b-ufuc}
        \bottomrule
    \end{tabular*}
    \caption{Glyphs generated by FVF-REF/B across the four test splits.
        Ground truth is shown in blue and generated glyphs in orange.
        Samples are not cherry-picked.}
    \label{fig:generalization_qualitative}
\end{figure}

Figure~\ref{fig:generalization_qualitative} shows glyphs generated by FlowVecFont-REF/B
on all four test splits.
Across the splits, Latin letters and symbols are generally well generated,
while Chinese characters often exhibit artifacts.
No clear difference in visual quality is apparent among the splits in these examples.

Table~\ref{tab:generalization_results} shows that CD remains similar across the splits,
while LPIPS and FID are higher on the unseen-font splits than on SFSC.
The failure rate is also highest on UFUC.
Nevertheless, performance on the unseen splits remains close to that on SFSC,
demonstrating the potential of reference conditioning to generalize vector glyph
generation to unseen fonts and characters.

\begin{table}[t]
    \centering
    \caption{VAE reconstruction and generation results on SFSC.
        Size denotes the total parameter count, including the shared frozen VAE.
        Fail.\ denotes the evaluation failure rate (\%).}
    \label{tab:conditioning_results}
    \small
    \begin{tabular*}{\columnwidth}{@{\extracolsep{\fill}}lrrrrr@{}}
        \toprule
        Model & Size & CD $\downarrow$ & LPIPS $\downarrow$ & FID $\downarrow$ & Fail.\ $\downarrow$ \\
        \midrule
        FVF-VAE & 14M & 0.0029 & 0.0045 & 0.09 & 0.00 \\
        \midrule
        FVF-CLS/S & 90M & 0.0194 & 0.1135 & 23.11 & 2.12 \\
        FVF-CLS/B & 230M & 0.0157 & 0.0853 & 11.86 & 1.17 \\
        FVF-CLS/L & 586M & 0.0136 & 0.0685 & 6.51 & 0.74 \\
        FVF-CLS/XL & 817M & 0.0131 & 0.0657 & 5.86 & 0.55 \\
        \midrule
        FVF-STR/S & 88M & 0.0279 & 0.1554 & 35.09 & 3.11 \\
        FVF-STR/B & 227M & 0.0267 & 0.1448 & 28.20 & 1.89 \\
        FVF-STR/L & 582M & 0.0209 & 0.1016 & 11.53 & 1.43 \\
        FVF-STR/XL & 813M & 0.0204 & 0.0991 & 10.62 & 1.18 \\
        \midrule
        FVF-REF/S & 47M & 0.0292 & 0.1611 & 37.03 & 3.80 \\
        FVF-REF/B & 144M & 0.0268 & 0.1414 & 26.87 & 2.67 \\
        FVF-REF/L & 471M & 0.0238 & 0.1166 & 14.80 & 1.94 \\
        FVF-REF/XL & 688M & 0.0232 & 0.1123 & 12.72 & 1.71 \\
        \bottomrule
    \end{tabular*}
\end{table}

\subsection{Comparison with Existing Methods}

We compare FlowVecFont with selected existing methods with publicly available weights:
the English DeepVecFont-v2 (DVF-v2) checkpoint, the DesigNet model,
and VecGlypher-27B with reference images.
We use our UFSC test split, so the target fonts are unseen during FlowVecFont training.
The common character set supported by the comparison methods consists of
uppercase and lowercase Latin letters.
We evaluate glyphs that satisfy all methods' representation constraints,
excluding DesigNet's fixed reference characters.
We use FlowVecFont-STR/XL to match the comparison setting,
where character identity and style references specify the target glyph.
Because the evaluation protocol differs from the previous experiments,
the scores are not directly comparable.

DeepVecFont-v2, VecGlypher, and FlowVecFont use the same four style references per target,
whereas DesigNet uses its fixed ``Hamburge'' references.
DeepVecFont-v2 selects one of 20 vector candidates using its predicted bitmap.

\begin{figure}[t]
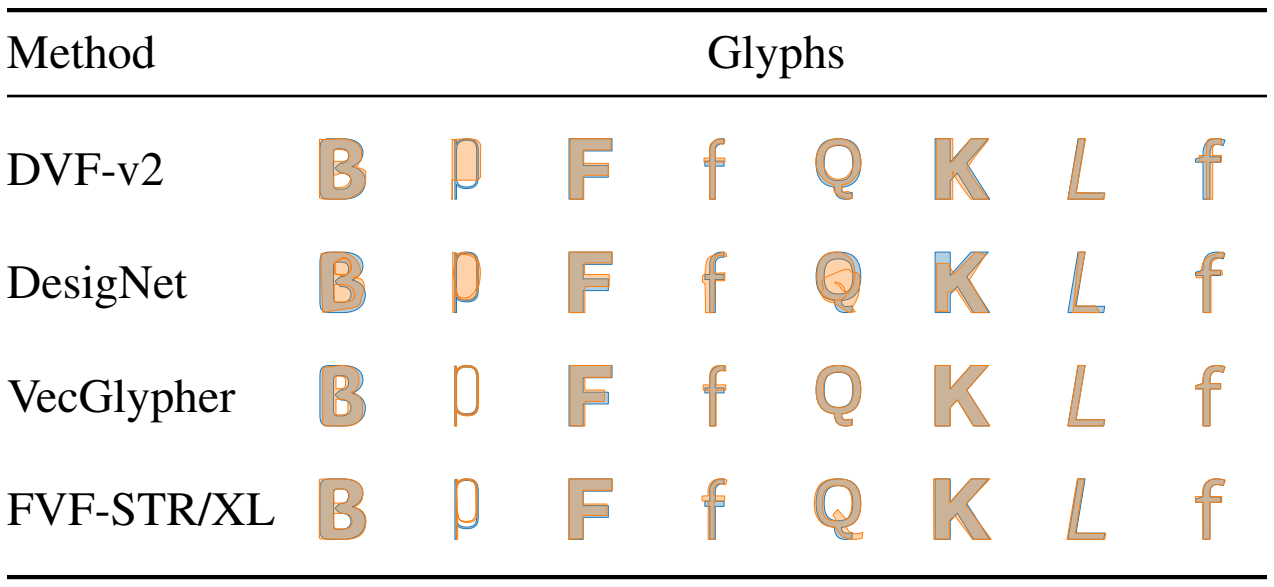

    \centering
    \small
    \setlength{\tabcolsep}{0pt}
    \begin{tabular*}{\linewidth}{@{\extracolsep{\fill}}l*{8}{c}@{}}
        \toprule
        Method & \multicolumn{8}{c@{}}{Glyphs} \\
        \midrule
        \glyphroweight{DVF-v2}{comparison-deepvecfont-v2}
        \glyphroweight{DesigNet}{comparison-designet}
        \glyphroweight{VecGlypher}{comparison-vecglypher}
        \glyphroweight{FVF-STR/XL}{comparison-flowvecfont}
        \bottomrule
    \end{tabular*}
    \caption{Qualitative comparison with existing methods on UFSC.
        Ground truth is shown in blue and generated glyphs in orange.
        Columns share the same targets; samples are not cherry-picked.}
    \label{fig:comparison_qualitative}
\end{figure}

Figure~\ref{fig:comparison_qualitative} shows that FlowVecFont generates glyphs
with outlines close to the targets in the displayed examples.
Table~\ref{tab:comparison_results} shows that FlowVecFont outperforms
DeepVecFont-v2 and DesigNet across all reported metrics, while VecGlypher achieves
the best quality scores.
Neither FlowVecFont nor VecGlypher has evaluation failures.
FlowVecFont has 800M parameters, substantially fewer than VecGlypher's 27B.
In addition to Google Fonts, VecGlypher is trained on an unreleased dataset of 39K commercial fonts,
and unlike FlowVecFont, its training data include some of the target fonts and glyphs.
Despite these differences, FlowVecFont achieves competitive quality,
demonstrating its effectiveness for vector glyph generation on unseen fonts.

\begin{table}[t]
    \centering
    \caption{FVF-REF/B results across the four test splits.}
    \label{tab:generalization_results}
    \small
    \begin{tabular*}{\columnwidth}{@{\extracolsep{\fill}}lrrrr@{}}
        \toprule
        Split & CD $\downarrow$ & LPIPS $\downarrow$ & FID $\downarrow$ & Fail.\ $\downarrow$ \\
        \midrule
        SFSC & 0.0268 & 0.1414 & 26.87 & 2.67 \\
        SFUC & 0.0264 & 0.1428 & 28.22 & 2.66 \\
        UFSC & 0.0273 & 0.1473 & 32.08 & 3.04 \\
        UFUC & 0.0268 & 0.1485 & 35.10 & 3.39 \\
        \bottomrule
    \end{tabular*}
\end{table}

\subsection{Limitations}

VecGlypher~\cite{huang2026vecglypher} uses family-level splitting with Latin-specific
filtering and deduplication to assess generalization across font families,
which is difficult to directly extend to all of Google Fonts.
A family-level split alone may also leave shared outlines across splits because
related or repackaged fonts, such as Noto variants, can belong to different families.
We therefore use a simpler face-level split,
in which different faces from the same family may belong to different partitions.
Different codepoints can represent the same or similar glyphs, such as similar Latin
and Cyrillic letters or CJK compatibility ideographs.
In some non-text fonts and TeX math fonts, codepoints do not represent their usual characters.
We do not account for these cases when splitting the dataset.
Glyphs in highly decorative fonts, such as Rubik Pixels, may exceed the sequence length limit.

\begin{table}[t]
    \centering
    \caption{Comparison with existing methods on 3,170 UFSC glyphs
        from 74 Google Fonts faces.
        Quality metrics are computed after normalizing each glyph using its own bounding box.}
    \label{tab:comparison_results}
    \small
    \begin{tabular*}{\columnwidth}{@{\extracolsep{\fill}}lrrrrr@{}}
        \toprule
        Method & Size & CD $\downarrow$ & LPIPS $\downarrow$ & FID $\downarrow$
        & Fail.\ $\downarrow$ \\
        \midrule
        DVF-v2
        & 121M & 0.0211 & 0.1627 & 24.99 & 0.41 \\
        DesigNet
        & 14M & 0.0269 & 0.2146 & 50.03 & 0.63 \\
        VecGlypher
        & 27B & \textbf{0.0129} & \textbf{0.0849} & \textbf{3.27} & \textbf{0.00} \\
        FVF-STR/XL
        & 813M & 0.0134 & 0.0917 & 4.86 & \textbf{0.00} \\
        \bottomrule
    \end{tabular*}
\end{table}

\section{Conclusion}

We presented FlowVecFont, a scalable model for pure vector font generation.
Its Transformer VAE maps glyphs to fixed-length latent tokens,
enabling flow matching while decoupling flow computation from outline length.
Experiments show that generation quality improves with model scale across conditioning
strategies, and that vector references enable generation for unseen fonts and characters.
On Latin glyphs, FlowVecFont outperforms DeepVecFont-v2 and DesigNet and achieves
competitive quality with the pretrained 27B-parameter VecGlypher.
These results show that scaling a simple model trained on large-scale vector data
offers a compelling alternative to relying on bitmap representations or external pretraining
for high-quality vector font generation.

\iftoggle{cvprfinal}{
    \section*{Acknowledgments}
}{
    \section*{LLM Usage Statement}
}

\iftoggle{cvprfinal}{
    This work was partly supported by JSPS KAKENHI Grant Number JP25K24744.
}{}

We used Codex (OpenAI) and Claude Code (Anthropic) interactively
for writing, coding, experiments, and figures, and reviewed and revised their outputs.
Some multi-step tasks, such as literature surveys and experiment monitoring,
were carried out autonomously.
The research ideas are our own, and the tools only helped refine them.
We take full responsibility for all content.

{
    \small
    \bibliographystyle{ieeenat_fullname}
    \bibliography{main}
}

\clearpage
\appendix
\setcounter{table}{0}
\setcounter{figure}{0}
\renewcommand{\thetable}{S\arabic{table}}
\renewcommand{\thefigure}{S\arabic{figure}}
\renewcommand{\theHtable}{supp.\arabic{table}}
\renewcommand{\theHfigure}{supp.\arabic{figure}}
\maketitlesupplementary

\begin{figure*}[!b]
    \centering
    \includegraphics[scale=0.6]{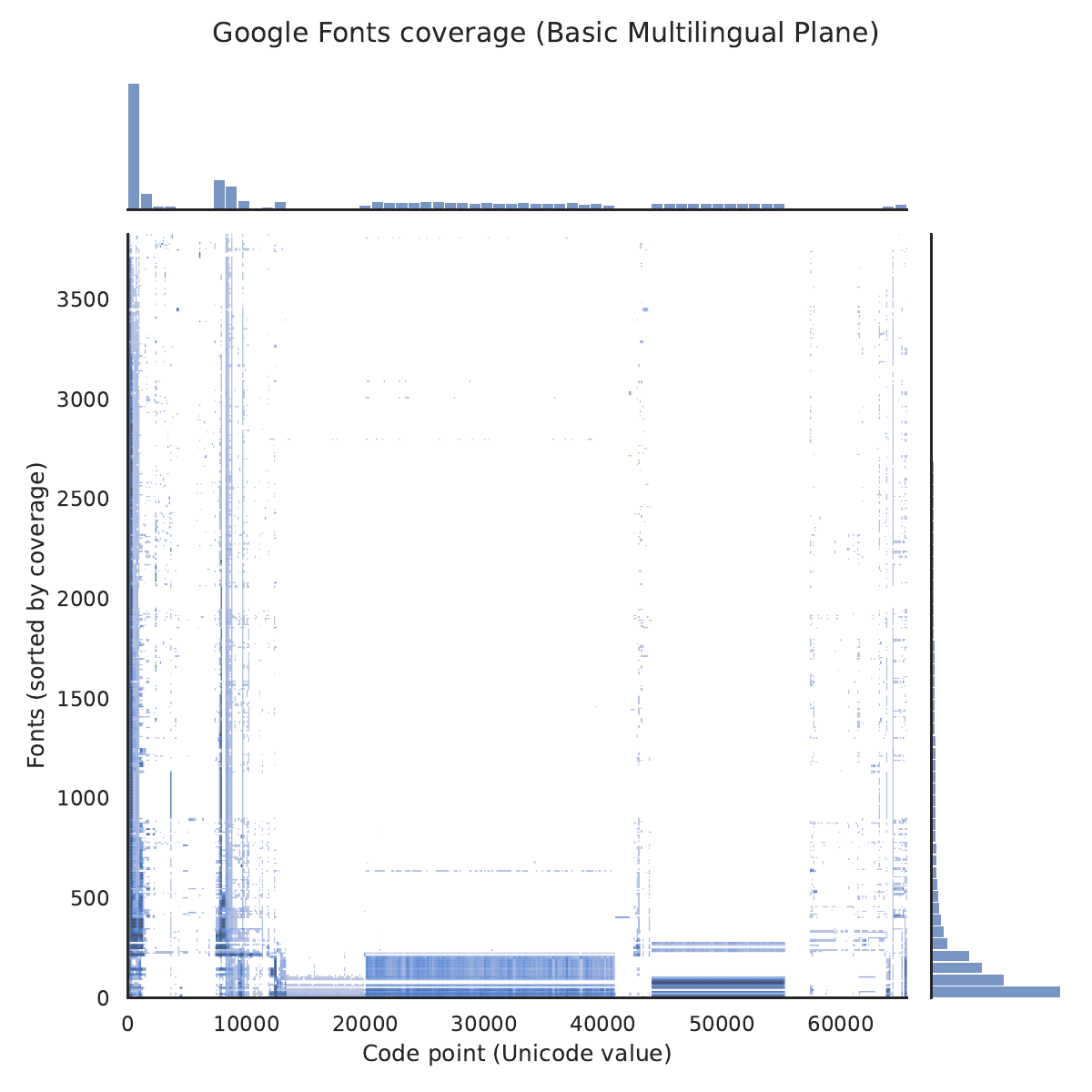}
    \caption{Google Fonts character coverage in the Basic Multilingual Plane (BMP).
        Colored bins indicate observed pairs of font faces and codepoints,
        with darker blue denoting more pairs per bin.}
    \label{fig:dataset_coverage}
\end{figure*}

\section{Dataset Statistics}

Figure~\ref{fig:dataset_coverage} shows character coverage in the full Google Fonts
collection used in this work, before preprocessing or sequence-length filtering.
Sample counts follow long-tailed distributions over both character and font labels.
The dense region at the lowest codepoints includes ASCII (Basic Latin) and Latin extensions,
which are supported by many fonts.
Other concentrations correspond to additional Latin characters and commonly used symbols.
The broad bands at higher codepoints correspond to CJK Unified Ideographs
and Hangul Syllables, both supported by relatively few fonts.

\begin{figure}[t]
    \centering
    \includegraphics[width=\linewidth]{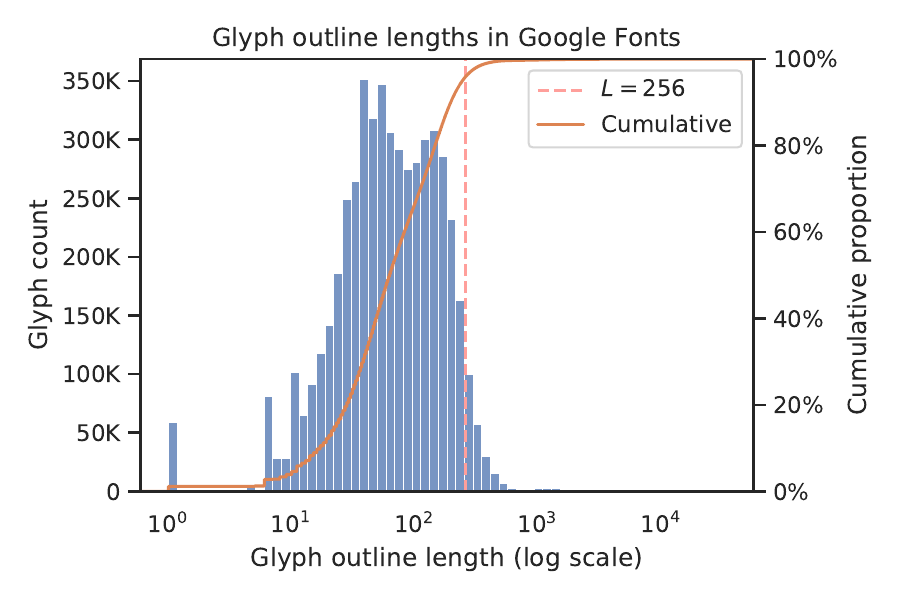}
    \caption{Original outline sequence lengths, including the End token,
        with cumulative proportions on the right axis.
        The dashed line marks $L=256$, which covers 95.9\% of all glyphs.}
    \label{fig:dataset_outline_lengths}
\end{figure}

Figure~\ref{fig:dataset_outline_lengths} shows the original outline sequence lengths
in the same collection, with variable fonts measured at their default axis settings.
The mean and median sequence lengths are 93.8 and 62 elements, respectively,
including the End token.
The maximum command sequence length $L=256$ covers 95.9\% of all glyphs.
The font families contributing the most glyphs above this limit are
Shippori Mincho B1, Chiron Sung HK, and Bpmf Zihi Kai Std.
Decorative fonts such as Rubik Pixels, Handjet, and Rubik 80s Fade also have
particularly long glyph sequences.

\section{Preprocessing}

VAE, CLS, STR, and REF use a common preprocessing framework, with transformations
and their strengths adapted to each model and conditioning scheme.
We present REF as a representative example because it uses both style and character
references.

\begin{figure}[t]
    \centering
    \includegraphics[scale=0.65]{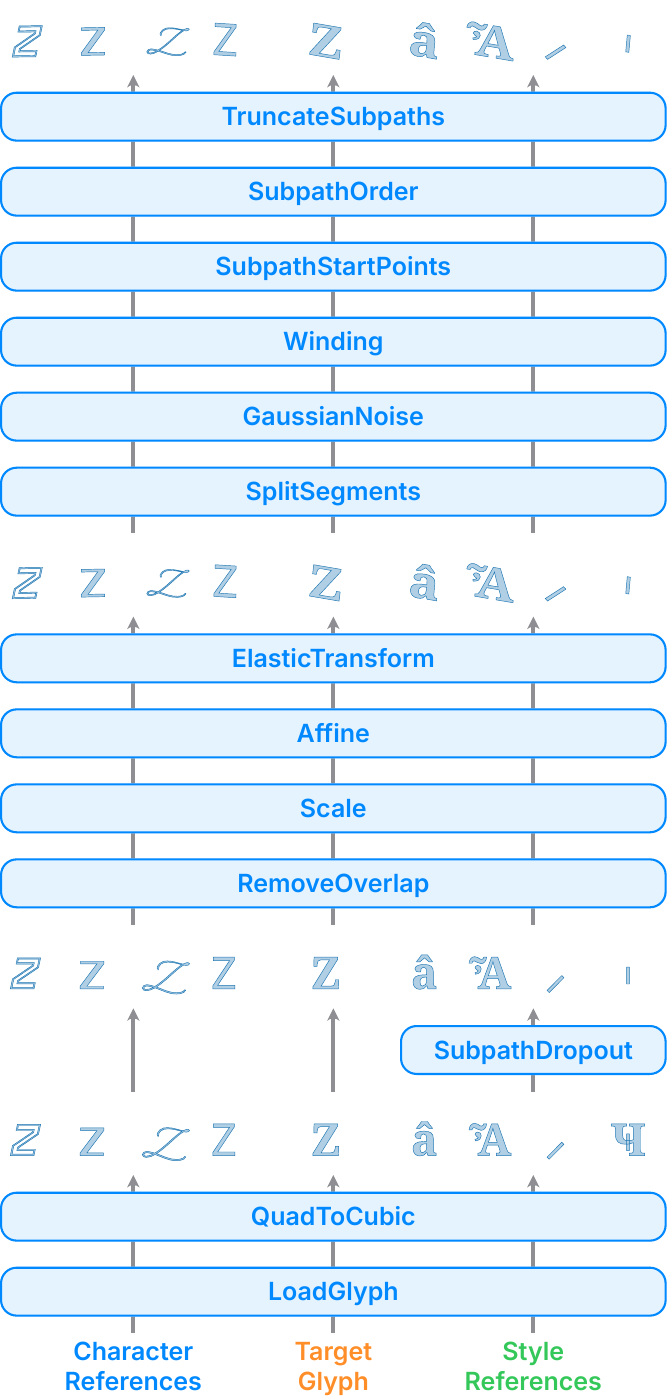}
    \caption{Preprocessing pipeline for conditional REF training.
        Blocks span the groups receiving each operation.}
    \label{fig:ref_train_preprocessing}
\end{figure}

Figure~\ref{fig:ref_train_preprocessing} illustrates the conditional REF training pipeline
for one target, four style references, and four character references.
For variable fonts, we vary glyph appearance by randomly sampling axis settings,
shared by the target and style references but sampled independently for each
character reference.
We then convert quadratic curves to cubic B\'ezier curves and merge compatible adjacent
segments to obtain a compact, consistent outline representation.

We apply subpath dropout independently to each style reference.
We then apply overlap processing, random scaling, affine transformations,
and elastic deformations to all glyphs.
For overlap processing, we choose with equal probability to leave the outline unchanged,
remove overlaps within randomly selected contour groups, or remove all overlaps.
The target and style references share the application decisions and transformation
parameters for these operations, while each character reference is processed independently.

The remaining random transformations are sampled independently for each glyph.
We apply segment splitting and Gaussian coordinate noise, with stronger perturbations
for character references.
We normalize winding, subpath starting points, and subpath order for the target and
style references.
For character references, each of these operations independently chooses between
normalization and randomization.
Finally, we randomly remove complete subpaths from sequences exceeding $L=256$
until they fit within the limit.

At evaluation time, preprocessing is deterministic for all four models.
We use the default axis settings for variable fonts and apply the same curve conversion
and merging steps.
We normalize winding, subpath starting points, and subpath order, then truncate
sequences at subpath boundaries if they exceed the length limit.

\section{Ablation Studies}

\subsection{Coordinate Reconstruction Loss}

We compare L1 and mean squared error (MSE) coordinate reconstruction losses in terms of
training stability and test reconstruction quality over 400K training steps.
The coordinate loss weights are 3 for L1 and 100 for MSE.
These weights were tuned separately in preliminary experiments for each objective.
We compute validation losses using exponential moving average (EMA) weights.

\begin{figure}[t]
    \centering
    \includegraphics[width=\linewidth]{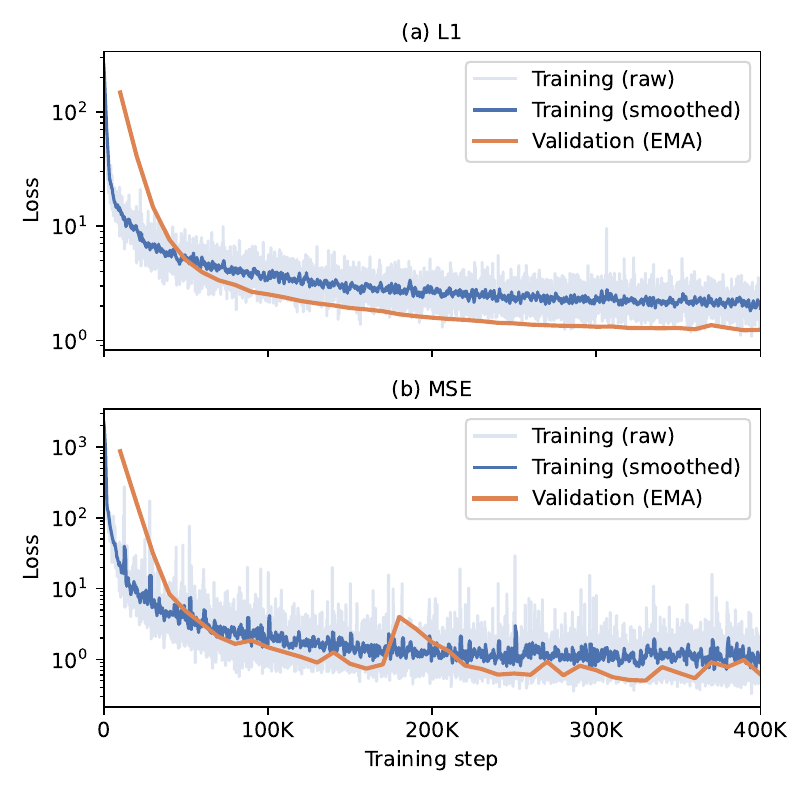}
    \caption{Training and validation losses for the L1 and MSE VAE variants.
        Faint blue lines show raw training losses before EMA, and solid blue lines show
        a trailing moving average over 1K training steps.
        Orange lines show validation losses computed every 10K steps using the EMA weights,
        without smoothing.
        Absolute loss values are not directly comparable because the objectives
        and their weights differ.}
    \label{fig:ablation_vae_loss_curve}
\end{figure}

Figure~\ref{fig:ablation_vae_loss_curve} shows that the L1 validation loss
decreases smoothly over most of training.
The MSE validation loss shows repeated spikes, especially around 180K steps.
It remains unstable even near the end of training.
EMA retains the influence of unstable training updates even after the training loss recovers.
Thus, additional training steps do not consistently reduce the MSE validation loss.

\begin{figure}[t]
    \centering
    \small
    \begin{tabular*}{\columnwidth}{@{\extracolsep{\fill}}llc@{}}
        \toprule
        Font & Codepoint & Outline \\
        \midrule
        Ubuntu Regular & \texttt{U+0041} &
        \glyphcenter{\includegraphics[width=0.49\linewidth]{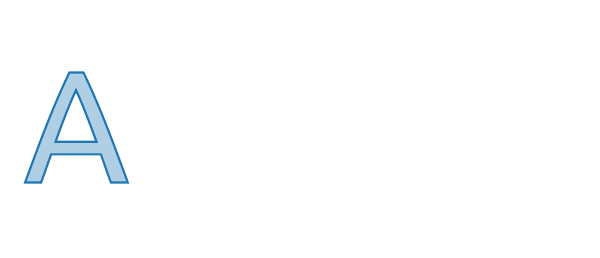}} \\
        \midrule
        Ubuntu Regular & \texttt{U+F200} &
        \glyphcenter{\includegraphics[width=0.49\linewidth]{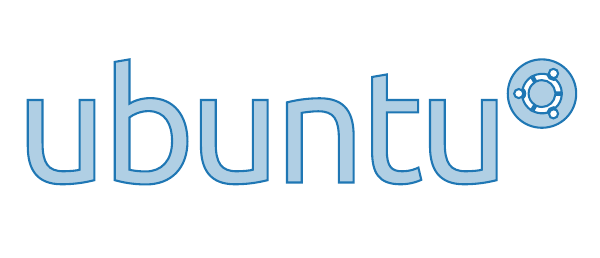}} \\
        \midrule
        MonteCarlo & \texttt{U+E16A} &
        \glyphcenter{\includegraphics[width=0.49\linewidth]{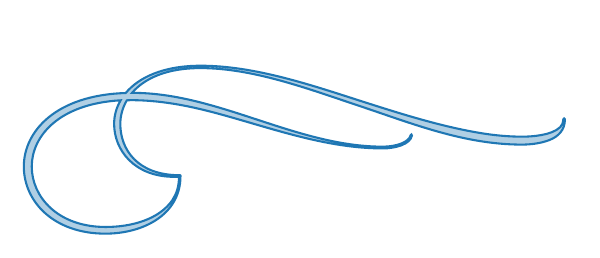}} \\
        \bottomrule
    \end{tabular*}
    \caption{Large-coordinate training glyphs found near loss spikes.
        The letter A provides a scale reference.}
    \label{fig:ablation_coordinate_outliers}
\end{figure}

Rare glyphs with unusually large coordinates contribute to this instability.
Figure~\ref{fig:ablation_coordinate_outliers} shows two examples:
the Ubuntu Regular wordmark (\texttt{U+F200})
and the MonteCarlo decorative swash (\texttt{U+E16A}).
Their maximum absolute coordinates before augmentation are 3.480 and 3.409 em,
respectively, compared with 0.693 em for the Ubuntu Regular letter A.
Our input coordinates are expressed in em units and are not normalized to each glyph's bounding box.
These unusually wide outlines therefore retain large coordinate values.
MSE applies a quadratic penalty to reconstruction errors,
making training sensitive to these glyphs.
L1 uses a linear penalty and is less sensitive to them.

Table~\ref{tab:ablation_vae_loss} shows that L1 also achieves lower CD, LPIPS, FID,
and evaluation failure rate than MSE on SFSC.
The improvement in both geometric and image-based metrics supports the use of L1
for coordinate reconstruction, alongside its more stable validation trajectory.

\begin{table}[t]
    \centering
    \caption{Test reconstruction quality on SFSC after 400K VAE training steps
        with L1 or MSE coordinate loss.
        Fail.\ denotes the failure rate (\%).}
    \label{tab:ablation_vae_loss}
    \small
    \begin{tabular*}{\columnwidth}{@{\extracolsep{\fill}}lrrrr@{}}
        \toprule
        Loss & CD $\downarrow$ & LPIPS $\downarrow$ & FID $\downarrow$ & Fail.\ $\downarrow$ \\
        \midrule
        L1 & \textbf{0.0029} & \textbf{0.0045} & \textbf{0.09} & \textbf{0.00} \\
        MSE & 0.0039 & 0.0093 & 0.30 & 0.01 \\
        \bottomrule
    \end{tabular*}
\end{table}

\subsection{VAE Data Augmentation}

\begin{table}[t]
    \centering
    \caption{VAE data augmentation on SFSC after 400K training steps.
        Fail.\ denotes the failure rate (\%).}
    \label{tab:ablation_vae_augmentation}
    \small
    \begin{tabular*}{\columnwidth}{@{\extracolsep{\fill}}lrrrr@{}}
        \toprule
        Augmentation & CD $\downarrow$ & LPIPS $\downarrow$ & FID $\downarrow$ & Fail.\ $\downarrow$ \\
        \midrule
        None & \textbf{0.0028} & \textbf{0.0039} & \textbf{0.05} & 0.023 \\
        w/o axis sampling & 0.0031 & 0.0053 & 0.11 & 0.002 \\
        Full & 0.0029 & 0.0045 & 0.09 & \textbf{0.000} \\
        \bottomrule
    \end{tabular*}
\end{table}

We compare three training settings: no augmentation (None);
augmentation with fixed variable-font axes (w/o axis sampling);
full augmentation with random axis sampling (Full).
Evaluation is performed without augmentation.

Table~\ref{tab:ablation_vae_augmentation} reports their test reconstruction quality.
Training without augmentation gives the lowest CD, LPIPS, and FID.
Augmentation reduces evaluation failures,
and no evaluation failures are observed with the full pipeline.
This suggests that augmentation improves reconstruction robustness.
Within the augmented training setting, random axis sampling also improves CD, LPIPS, and FID
compared with fixed axes.

Pipelines based on offline preprocessing, such as DeepVecFont-v2~\cite{wang2023deepvecfont}
and DesigNet~\cite{guijavaliente2026designet}, use extracted glyph representations
without the font's variation data.
This limits training to a finite set of axis positions.
\iftoggle{cvprfinal}{TorchFont, our library,}{Our library}
loads outlines directly from font files on the fly.
This allows new axis positions to be sampled randomly from a continuous space during training.
The improvement from axis sampling demonstrates the benefit of this capability.

\subsection{Number of Latent Tokens}

We compare 16, 32, and 64 latent tokens for VAE reconstruction and font generation.
Each VAE and its corresponding FVF-CLS/B model are trained for 400K steps.

Table~\ref{tab:ablation_tokens} reports reconstruction and generation quality on SFSC.
Using 64 tokens gives the lowest CD, LPIPS, and FID for both models.
VAE reconstruction improves substantially from 16 to 32 tokens.
In contrast, increasing the token count from 32 to 64 yields only small further improvements.
FVF-CLS/B continues to improve at 64 tokens, but longer latent sequences also increase
the computational cost per training step.
Since all models are trained for the same number of steps, these gains may partly reflect
the additional computation.
We use 32 tokens because they provide sufficient reconstruction quality
while requiring less computation than 64 tokens.

\begin{table}[t]
    \centering
    \caption{Effect of the number of latent tokens on FVF-VAE reconstruction
        and FVF-CLS/B generation on SFSC.
        In this ablation, FVF-CLS/B is trained for 400K steps instead of the 1M steps
        used in the main experiments.
        Fail.\ denotes the failure rate (\%).}
    \label{tab:ablation_tokens}
    \small
    \setlength{\tabcolsep}{3pt}
    \begin{tabular*}{\columnwidth}{@{\extracolsep{\fill}}llrrrr@{}}
        \toprule
        Model & $K$ & CD $\downarrow$ & LPIPS $\downarrow$ & FID $\downarrow$ & Fail.\ $\downarrow$ \\
        \midrule
        \multirow{3}{*}{FVF-VAE} & 16 & 0.0041 & 0.0166 & 1.38 & 0.121 \\
        & 32 & 0.0029 & 0.0045 & 0.09 & \textbf{0.000} \\
        & 64 & \textbf{0.0025} & \textbf{0.0030} & \textbf{0.04} & 0.002 \\
        \midrule
        \multirow{3}{*}{FVF-CLS/B} & 16 & 0.0191 & 0.1168 & 26.79 & 2.21 \\
        & 32 & 0.0183 & 0.1018 & 16.40 & 1.71 \\
        & 64 & \textbf{0.0171} & \textbf{0.0861} & \textbf{7.22} & \textbf{0.44} \\
        \bottomrule
    \end{tabular*}
\end{table}

\subsection{Guidance Strength and Euler Steps}

We evaluate the effects of classifier-free guidance (CFG) strength and the number of Euler steps on SFSC
using the same CLS/B checkpoint trained for 1M steps.
We compare CFG scales of $\{1.0,1.5,2.0,3.0,4.0,5.0\}$ with 32 Euler steps,
and $\{8,16,32,64\}$ Euler steps with CFG scale 3.0.

Table~\ref{tab:ablation_cfg} shows that increasing the CFG scale from 1.0 to 3.0
improves CD, LPIPS, and FID.
However, increasing the scale to 4.0 or 5.0 makes all three metrics worse.
Therefore, scale 3.0 gives the best generation quality among the scales we compare.

\begin{table}[t]
    \centering
    \caption{Guidance-strength ablation for the 1M-step CLS/B checkpoint on SFSC
        with 32 Euler steps.
        Fail.\ denotes the failure rate (\%).}
    \label{tab:ablation_cfg}
    \small
    \begin{tabular*}{\columnwidth}{@{\extracolsep{\fill}}lrrrr@{}}
        \toprule
        CFG scale & CD $\downarrow$ & LPIPS $\downarrow$ & FID $\downarrow$ & Fail.\ $\downarrow$ \\
        \midrule
        1.0 & 0.0207 & 0.1146 & 18.26 & 1.26 \\
        1.5 & 0.0167 & 0.0919 & 13.07 & 1.16 \\
        2.0 & 0.0159 & 0.0869 & 12.02 & \textbf{1.14} \\
        3.0 & \textbf{0.0157} & \textbf{0.0853} & \textbf{11.87} & 1.21 \\
        4.0 & 0.0158 & 0.0857 & 12.17 & 1.20 \\
        5.0 & 0.0159 & 0.0864 & 12.51 & 1.20 \\
        \bottomrule
    \end{tabular*}
\end{table}

Table~\ref{tab:ablation_steps} shows that 32 Euler steps give the best generation quality
among the step counts we compare.
Increasing the number of steps to 64 makes CD, LPIPS, and FID slightly worse,
but slightly lowers the failure rate.
Based on these results, we use CFG scale 3.0 and 32 Euler steps for inference.

\begin{table}[t]
    \centering
    \caption{Euler-step ablation for the 1M-step CLS/B checkpoint on SFSC.
        All variants use CFG scale 3.0.
        Fail.\ denotes the failure rate (\%).
        Best values before rounding are in bold.}
    \label{tab:ablation_steps}
    \small
    \begin{tabular*}{\columnwidth}{@{\extracolsep{\fill}}lrrrr@{}}
        \toprule
        Euler steps & CD $\downarrow$ & LPIPS $\downarrow$ & FID $\downarrow$ & Fail.\ $\downarrow$ \\
        \midrule
        8 & 0.0163 & 0.0933 & 20.63 & 1.79 \\
        16 & 0.0157 & 0.0859 & 13.01 & 1.33 \\
        32 & \textbf{0.0157} & \textbf{0.0853} & \textbf{11.87} & 1.21 \\
        64 & 0.0157 & 0.0856 & 12.05 & \textbf{1.18} \\
        \bottomrule
    \end{tabular*}
\end{table}

\section{Additional Results}

Table~\ref{tab:full_results} provides the full reconstruction and generation results
across model sizes and test splits.
Alongside symmetric Chamfer distance (CD)~\cite{barrow1977parametric},
learned perceptual image patch similarity (LPIPS)~\cite{zhang2018perceptual},
Fr\'echet Inception Distance (FID)~\cite{heusel2017gans}, and the evaluation failure rate,
we report mean absolute pixel error (L1) and structural similarity
(SSIM)~\cite{wang2004image}.

The VAE maintains low reconstruction error across all four splits.
For generation, L1 decreases and SSIM increases with model size for every evaluated
conditioning strategy and split, consistent with the scaling trends in the main paper.

\begin{figure*}[p]
    \centering
    \includegraphics[width=0.49\textwidth]{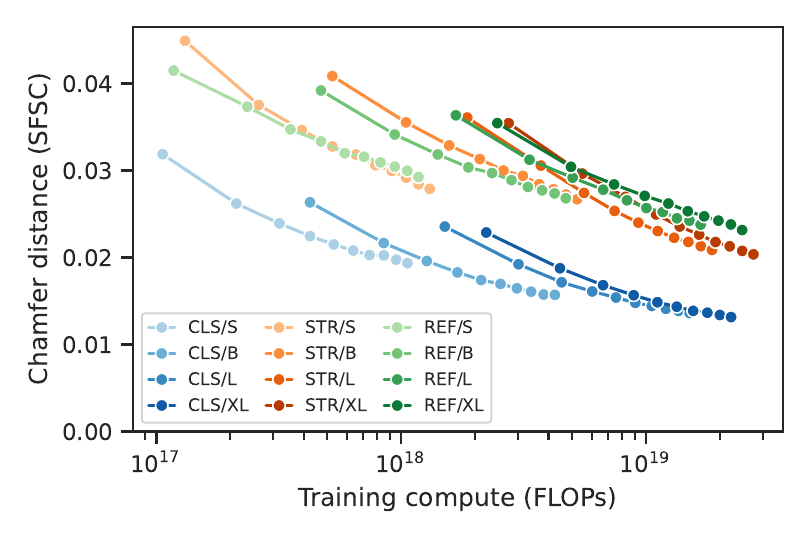}
    \hfill
    \includegraphics[width=0.49\textwidth]{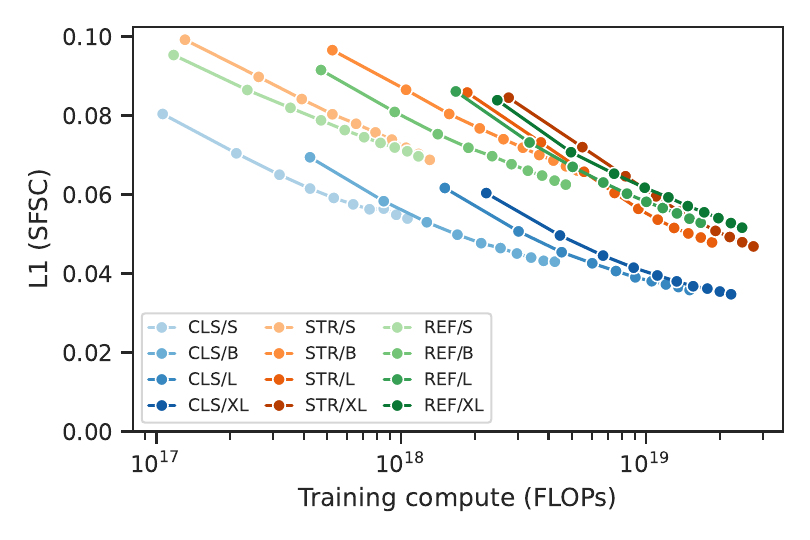}
    \\[1ex]
    \includegraphics[width=0.49\textwidth]{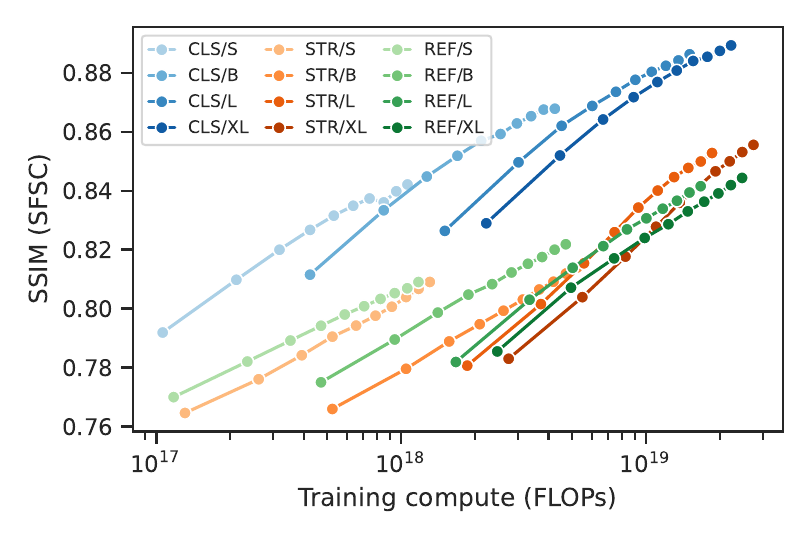}
    \hfill
    \includegraphics[width=0.49\textwidth]{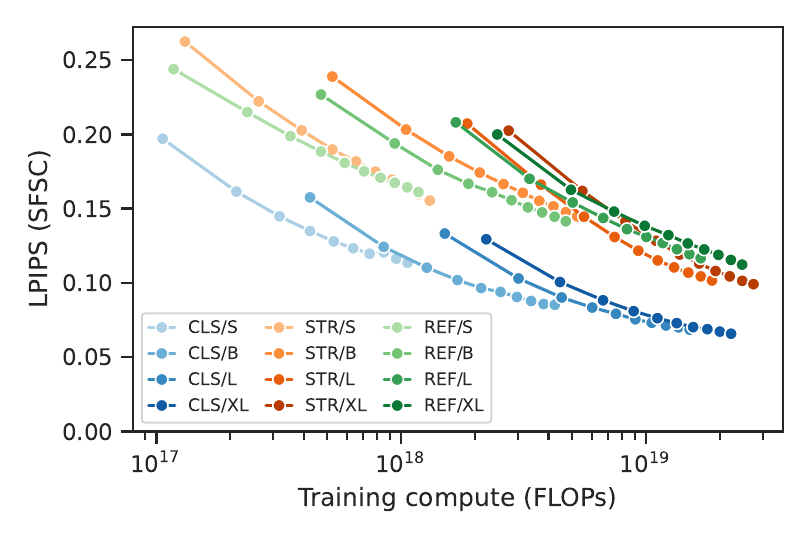}
    \\[1ex]
    \includegraphics[width=0.49\textwidth]{figures/scaling/fid.pdf}
    \hfill
    \includegraphics[width=0.49\textwidth]{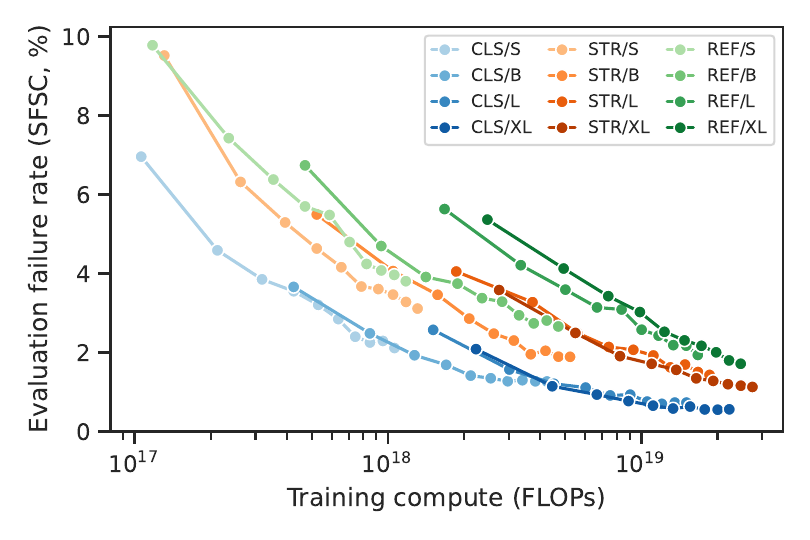}
    \caption{Chamfer distance, L1, SSIM, LPIPS, FID, and evaluation failure rate on SFSC
        versus training compute.
        Each point corresponds to a checkpoint saved every 100K steps.}
    \label{fig:scaling_full}
\end{figure*}

Figure~\ref{fig:scaling_full} extends the training compute analysis to all six metrics
on SFSC, using checkpoints saved every 100K training steps.

\begin{figure*}[p]
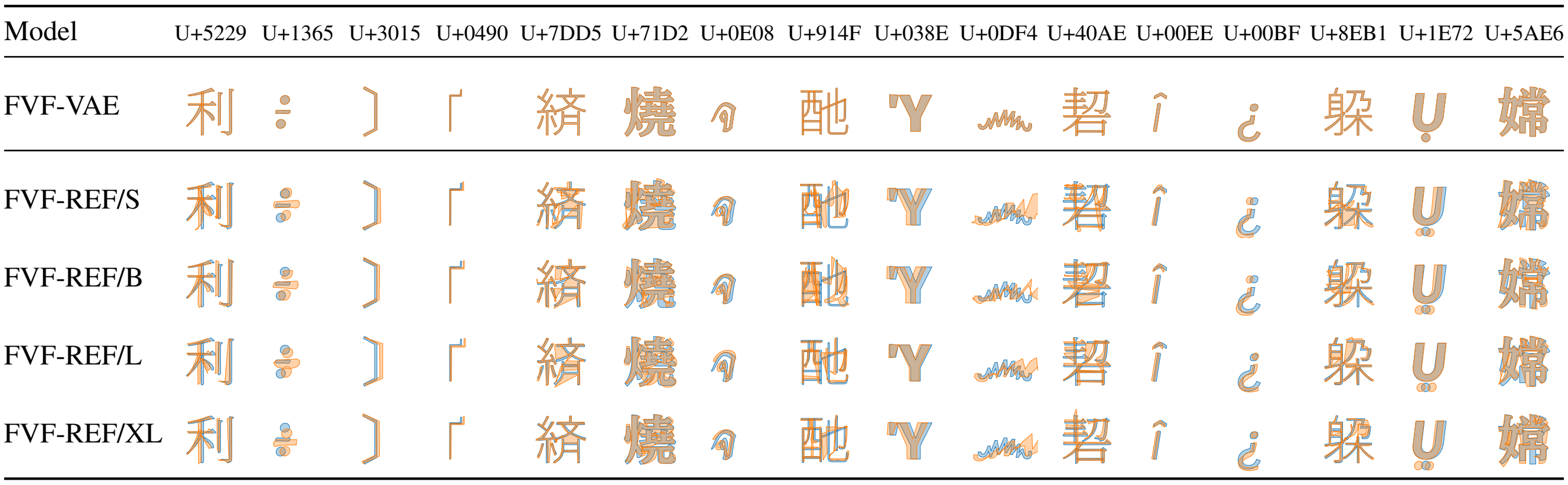

    \centering
    \small
    \setlength{\tabcolsep}{0pt}
    \begin{tabular*}{\textwidth}{@{\extracolsep{\fill}}l*{16}{c}@{}}
        \toprule
        Model \cp{5229}\cp{1365}\cp{3015}\cp{0490}\cp{7DD5}\cp{71D2}\cp{0E08}\cp{914F}\cp{038E}\cp{0DF4}\cp{40AE}\cp{00EE}\cp{00BF}\cp{8EB1}\cp{1E72}\cp{5AE6} \\
        \midrule
        \glyphrow{FVF-VAE}{vae-ufuc}
        \midrule
        \glyphrow{FVF-REF/S}{ref-s-ufuc}
        \glyphrow{FVF-REF/B}{ref-b-ufuc}
        \glyphrow{FVF-REF/L}{ref-l-ufuc}
        \glyphrow{FVF-REF/XL}{ref-xl-ufuc}
        \bottomrule
    \end{tabular*}
    \caption{Qualitative results on UFUC.
        Ground-truth glyphs (blue) are overlaid with VAE reconstructions or generated glyphs (orange).
        Each column shows the same target glyph, labeled with its Unicode codepoint, and samples are not cherry-picked.}
    \label{fig:qualitative_ufuc}
\end{figure*}

Figures~\ref{fig:qualitative_sfsc}, \ref{fig:qualitative_sfuc}, \ref{fig:qualitative_ufsc},
and~\ref{fig:qualitative_ufuc} show VAE reconstructions and generated glyphs on
SFSC, SFUC, UFSC, and UFUC, respectively, covering all evaluated conditioning strategies
and model sizes.
Each figure includes all 16 saved examples for its split.
Ground-truth outlines are shown in blue and reconstructions or generated outlines in
orange, allowing direct comparison of contour alignment and local artifacts.

Table~\ref{tab:full_comparison_results} adds L1 and SSIM to the comparison with
existing methods on 3,170 UFSC glyphs from 74 font faces.
These metrics give the same ranking as CD, LPIPS, and FID:
FVF-STR/XL outperforms DeepVecFont-v2 and DesigNet, while VecGlypher achieves the
best scores.
As in the main paper, each glyph is normalized using its own bounding box for this
comparison, so these scores are not directly comparable to those in
Table~\ref{tab:full_results}.

\begin{figure*}[p]
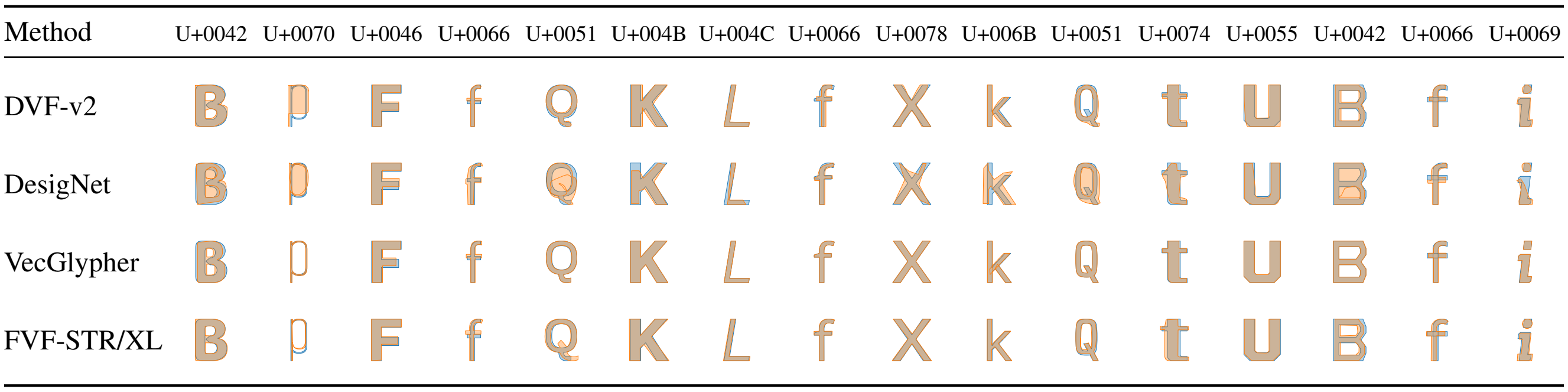

    \centering
    \small
    \setlength{\tabcolsep}{0pt}
    \begin{tabular*}{\textwidth}{@{\extracolsep{\fill}}l*{16}{c}@{}}
        \toprule
        Method \cp{0042}\cp{0070}\cp{0046}\cp{0066}\cp{0051}\cp{004B}\cp{004C}\cp{0066}\cp{0078}\cp{006B}\cp{0051}\cp{0074}\cp{0055}\cp{0042}\cp{0066}\cp{0069} \\
        \midrule
        \glyphrow{DVF-v2}{comparison-deepvecfont-v2}
        \glyphrow{DesigNet}{comparison-designet}
        \glyphrow{VecGlypher}{comparison-vecglypher}
        \glyphrow{FVF-STR/XL}{comparison-flowvecfont}
        \bottomrule
    \end{tabular*}
    \caption{Full qualitative comparison with existing methods on UFSC.
        FVF-STR/XL is our proposed method.
        Ground-truth glyphs (blue) are overlaid with generated glyphs (orange).
        Each column shows the same target glyph, labeled with its Unicode codepoint.
        All 16 saved examples are shown in ascending font-ID and codepoint order.}
    \label{fig:qualitative_comparison_full}
\end{figure*}

Figure~\ref{fig:qualitative_comparison_full} complements this quantitative comparison
with all 16 saved examples from the same evaluation.

\begin{table*}[p]
    \centering
    \caption{Full test results for VAE reconstruction and font generation.
        CD denotes Chamfer distance.
        Size denotes the total parameter count, including the shared frozen VAE.
        Fail.\ denotes the evaluation failure rate (\%).
        Quality metrics are computed on successfully evaluated glyphs.}
    \label{tab:full_results}
    \small
    \begin{tabular*}{\textwidth}{@{\extracolsep{\fill}}lrlrrrrrr@{}}
        \toprule
        Model & Size & Split & CD $\downarrow$ & L1 $\downarrow$ & SSIM $\uparrow$
        & LPIPS $\downarrow$ & FID $\downarrow$ & Fail.\ $\downarrow$ \\
        \midrule
        \multirow{4}{*}{FVF-VAE} & \multirow{4}{*}{14M} & SFSC & 0.0029 & 0.0031 & 0.9947 & 0.0045 & 0.09 & 0.00 \\
        & & SFUC & 0.0029 & 0.0031 & 0.9947 & 0.0045 & 0.07 & 0.00 \\
        & & UFSC & 0.0031 & 0.0034 & 0.9945 & 0.0047 & 0.06 & 0.00 \\
        & & UFUC & 0.0031 & 0.0034 & 0.9945 & 0.0047 & 0.16 & 0.00 \\
        \midrule
        FVF-CLS/S & 90M & SFSC & 0.0194 & 0.0539 & 0.8422 & 0.1135 & 23.11 & 2.12 \\
        \addlinespace[0.3em]
        FVF-CLS/B & 230M & SFSC & 0.0157 & 0.0431 & 0.8677 & 0.0853 & 11.86 & 1.17 \\
        \addlinespace[0.3em]
        FVF-CLS/L & 586M & SFSC & 0.0136 & 0.0358 & 0.8863 & 0.0685 & 6.51 & 0.74 \\
        \addlinespace[0.3em]
        FVF-CLS/XL & 817M & SFSC & 0.0131 & 0.0347 & 0.8894 & 0.0657 & 5.86 & 0.55 \\
        \midrule
        \multirow{2}{*}{FVF-STR/S} & \multirow{2}{*}{88M} & SFSC & 0.0279 & 0.0688 & 0.8091 & 0.1554 & 35.09 & 3.11 \\
        & & UFSC & 0.0285 & 0.0734 & 0.8000 & 0.1628 & 40.43 & 3.85 \\
        \addlinespace[0.3em]
        \multirow{2}{*}{FVF-STR/B} & \multirow{2}{*}{227M} & SFSC & 0.0267 & 0.0661 & 0.8140 & 0.1448 & 28.20 & 1.89 \\
        & & UFSC & 0.0277 & 0.0714 & 0.8031 & 0.1546 & 32.48 & 2.20 \\
        \addlinespace[0.3em]
        \multirow{2}{*}{FVF-STR/L} & \multirow{2}{*}{582M} & SFSC & 0.0209 & 0.0479 & 0.8528 & 0.1016 & 11.53 & 1.43 \\
        & & UFSC & 0.0217 & 0.0511 & 0.8458 & 0.1066 & 11.68 & 1.89 \\
        \addlinespace[0.3em]
        \multirow{2}{*}{FVF-STR/XL} & \multirow{2}{*}{813M} & SFSC & 0.0204 & 0.0469 & 0.8557 & 0.0991 & 10.62 & 1.18 \\
        & & UFSC & 0.0212 & 0.0502 & 0.8480 & 0.1036 & 10.72 & 1.50 \\
        \midrule
        \multirow{4}{*}{FVF-REF/S} & \multirow{4}{*}{47M} & SFSC & 0.0292 & 0.0697 & 0.8091 & 0.1611 & 37.03 & 3.80 \\
        & & SFUC & 0.0286 & 0.0708 & 0.8065 & 0.1627 & 38.98 & 3.66 \\
        & & UFSC & 0.0294 & 0.0728 & 0.8042 & 0.1663 & 43.63 & 4.32 \\
        & & UFUC & 0.0286 & 0.0739 & 0.8010 & 0.1678 & 48.22 & 4.19 \\
        \addlinespace[0.3em]
        \multirow{4}{*}{FVF-REF/B} & \multirow{4}{*}{144M} & SFSC & 0.0268 & 0.0625 & 0.8219 & 0.1414 & 26.87 & 2.67 \\
        & & SFUC & 0.0264 & 0.0634 & 0.8196 & 0.1428 & 28.22 & 2.66 \\
        & & UFSC & 0.0273 & 0.0661 & 0.8155 & 0.1473 & 32.08 & 3.04 \\
        & & UFUC & 0.0268 & 0.0669 & 0.8128 & 0.1485 & 35.10 & 3.39 \\
        \addlinespace[0.3em]
        \multirow{4}{*}{FVF-REF/L} & \multirow{4}{*}{471M} & SFSC & 0.0238 & 0.0529 & 0.8416 & 0.1166 & 14.80 & 1.94 \\
        & & SFUC & 0.0236 & 0.0536 & 0.8394 & 0.1178 & 15.34 & 2.07 \\
        & & UFSC & 0.0244 & 0.0560 & 0.8353 & 0.1216 & 16.59 & 2.34 \\
        & & UFUC & 0.0240 & 0.0572 & 0.8319 & 0.1233 & 18.86 & 2.28 \\
        \addlinespace[0.3em]
        \multirow{4}{*}{FVF-REF/XL} & \multirow{4}{*}{688M} & SFSC & 0.0232 & 0.0516 & 0.8444 & 0.1123 & 12.72 & 1.71 \\
        & & SFUC & 0.0232 & 0.0526 & 0.8418 & 0.1140 & 13.11 & 1.79 \\
        & & UFSC & 0.0238 & 0.0547 & 0.8384 & 0.1172 & 13.84 & 2.04 \\
        & & UFUC & 0.0235 & 0.0558 & 0.8351 & 0.1186 & 16.28 & 2.11 \\
        \bottomrule
    \end{tabular*}
\end{table*}

\begin{figure*}[p]
    \centering
    \small
    \setlength{\tabcolsep}{0pt}
    \begin{tabular*}{\textwidth}{@{\extracolsep{\fill}}l*{16}{c}@{}}
        \toprule
        Model \cp{88C5}\cp{1DF18}\cp{81A0}\cp{715C}\cp{00CB}\cp{219F}\cp{0415}\cp{C038}\cp{2A91}\cp{2569}\cp{00A4}\cp{02C7}\cp{5E73}\cp{1E08}\cp{55B6}\cp{1EF5} \\
        \midrule
        \glyphrow{FVF-VAE}{vae-sfsc}
        \midrule
        \glyphrow{FVF-CLS/S}{cls-s-sfsc}
        \glyphrow{FVF-CLS/B}{cls-b-sfsc}
        \glyphrow{FVF-CLS/L}{cls-l-sfsc}
        \glyphrow{FVF-CLS/XL}{cls-xl-sfsc}
        \midrule
        \glyphrow{FVF-STR/S}{str-s-sfsc}
        \glyphrow{FVF-STR/B}{str-b-sfsc}
        \glyphrow{FVF-STR/L}{str-l-sfsc}
        \glyphrow{FVF-STR/XL}{str-xl-sfsc}
        \midrule
        \glyphrow{FVF-REF/S}{ref-s-sfsc}
        \glyphrow{FVF-REF/B}{ref-b-sfsc}
        \glyphrow{FVF-REF/L}{ref-l-sfsc}
        \glyphrow{FVF-REF/XL}{ref-xl-sfsc}
        \bottomrule
    \end{tabular*}
    \caption{Qualitative results on SFSC.
        Ground-truth glyphs (blue) are overlaid with VAE reconstructions or generated glyphs (orange).
        Each column shows the same target glyph, labeled with its Unicode codepoint, and samples are not cherry-picked.}
    \label{fig:qualitative_sfsc}
\end{figure*}

\begin{figure*}[p]
    \centering
    \small
    \setlength{\tabcolsep}{0pt}
    \begin{tabular*}{\textwidth}{@{\extracolsep{\fill}}l*{16}{c}@{}}
        \toprule
        Model \cp{207E}\cp{CE49}\cp{5099}\cp{1EB7}\cp{53F6}\cp{C185}\cp{25CA}\cp{5094}\cp{00DD}\cp{B294}\cp{304E}\cp{6AAC}\cp{CCF7}\cp{00DD}\cp{B871}\cp{1E51} \\
        \midrule
        \glyphrow{FVF-VAE}{vae-sfuc}
        \midrule
        \glyphrow{FVF-REF/S}{ref-s-sfuc}
        \glyphrow{FVF-REF/B}{ref-b-sfuc}
        \glyphrow{FVF-REF/L}{ref-l-sfuc}
        \glyphrow{FVF-REF/XL}{ref-xl-sfuc}
        \bottomrule
    \end{tabular*}
    \caption{Qualitative results on SFUC.
        Ground-truth glyphs (blue) are overlaid with VAE reconstructions or generated glyphs (orange).
        Each column shows the same target glyph, labeled with its Unicode codepoint, and samples are not cherry-picked.}
    \label{fig:qualitative_sfuc}
\end{figure*}

\begin{figure*}[p]
    \centering
    \small
    \setlength{\tabcolsep}{0pt}
    \begin{tabular*}{\textwidth}{@{\extracolsep{\fill}}l*{16}{c}@{}}
        \toprule
        Model \cp{9B3C}\cp{00A1}\cp{00D3}\cp{4ACA}\cp{3641}\cp{213A}\cp{0022}\cp{8CB6}\cp{0132}\cp{7B18}\cp{1E90}\cp{9E11}\cp{816F}\cp{012C}\cp{C220}\cp{2135} \\
        \midrule
        \glyphrow{FVF-VAE}{vae-ufsc}
        \midrule
        \glyphrow{FVF-STR/S}{str-s-ufsc}
        \glyphrow{FVF-STR/B}{str-b-ufsc}
        \glyphrow{FVF-STR/L}{str-l-ufsc}
        \glyphrow{FVF-STR/XL}{str-xl-ufsc}
        \midrule
        \glyphrow{FVF-REF/S}{ref-s-ufsc}
        \glyphrow{FVF-REF/B}{ref-b-ufsc}
        \glyphrow{FVF-REF/L}{ref-l-ufsc}
        \glyphrow{FVF-REF/XL}{ref-xl-ufsc}
        \bottomrule
    \end{tabular*}
    \caption{Qualitative results on UFSC.
        Ground-truth glyphs (blue) are overlaid with VAE reconstructions or generated glyphs (orange).
        Each column shows the same target glyph, labeled with its Unicode codepoint, and samples are not cherry-picked.}
    \label{fig:qualitative_ufsc}
\end{figure*}

\clearpage

\begin{table*}[p]
    \centering
    \caption{Full comparison results on 3,170 UFSC glyphs from 74 Google Fonts faces.
        CD denotes Chamfer distance.
        Quality metrics are computed after normalizing each glyph using its own bounding box.
        Size denotes the parameter count, including frozen components used for inference.
        Fail.\ denotes the evaluation failure rate (\%).
        Quality metrics exclude failed samples.
        FVF-STR/XL is our proposed method.
        Best results are in bold.}
    \label{tab:full_comparison_results}
    \small
    \begin{tabular*}{\textwidth}{@{\extracolsep{\fill}}lrrrrrrr@{}}
        \toprule
        Method & Size & CD $\downarrow$ & L1 $\downarrow$ & SSIM $\uparrow$
        & LPIPS $\downarrow$ & FID $\downarrow$ & Fail.\ $\downarrow$ \\
        \midrule
        DVF-v2
        & 121M & 0.0211 & 0.0754 & 0.7793 & 0.1627 & 24.99 & 0.41 \\
        DesigNet
        & 14M & 0.0269 & 0.0989 & 0.7448 & 0.2146 & 50.03 & 0.63 \\
        VecGlypher
        & 27B & \textbf{0.0129} & \textbf{0.0464} & \textbf{0.8502}
        & \textbf{0.0849} & \textbf{3.27} & \textbf{0.00} \\
        FVF-STR/XL & 813M & 0.0134 & 0.0482 & 0.8353 & 0.0917
        & 4.86 & \textbf{0.00} \\
        \bottomrule
    \end{tabular*}
\end{table*}

\clearpage

\end{document}